\documentclass[10pt,journal]{IEEEtran}

\usepackage{amsmath}
\usepackage{amssymb}
\usepackage{xcolor}

\definecolor{aj}{rgb}{0.8, 0.33, 0.0}

\definecolor{ms}{rgb}{0.1, 0.5, 0.5}

\usepackage[utf8]{inputenc}
\ifCLASSOPTIONcompsoc
  \usepackage[nocompress]{cite}
\else
  \usepackage{cite}
\fi

\ifCLASSINFOpdf
\else

\fi

\usepackage{amsmath}

\usepackage{acronym}

\usepackage{algorithm,algorithmic}

\usepackage{array}
\usepackage{mdwmath}
\usepackage{mdwtab}
\usepackage{booktabs}

\usepackage{eqparbox}

\ifCLASSOPTIONcompsoc
  \usepackage[caption=false,font=footnotesize,labelfont=sf,textfont=sf]{subfig}
\else
  \usepackage[caption=false,font=footnotesize]{subfig}
\fi

\usepackage{fixltx2e}
\usepackage{stfloats}

\usepackage{graphicx}

\ifCLASSOPTIONcaptionsoff
  \usepackage[nomarkers]{endfloat}
 \let\MYoriglatexcaption\caption
 \renewcommand{\caption}[2][\relax]{\MYoriglatexcaption[#2]{#2}}
\fi

\usepackage{url}
\usepackage{xcolor}
  
\DeclareMathOperator*{\argmin}{arg\, min} 
\renewcommand\IEEEkeywordsname{Index Terms}

\ifCLASSINFOpdf
  \usepackage[pdftex]{thumbpdf}
\else
  \usepackage[dvips]{thumbpdf}
\fi

\begin{document}

\title{SPI-GAN: Towards Single-Pixel Imaging through Generative Adversarial Network}
\author{Nazmul Karim and Nazanin Rahnavard\\

\IEEEcompsocitemizethanks{\IEEEcompsocthanksitem Nazmul Karim and Nazanin Rahnavard are with the Department
of Electrical Engineering, University of Central Florida, Orlando,
FL, 32816. (e-mail: nazmul.karim18@knights.ucf.edu; nazanin.rahnavard@ucf.edu)
}
}
\maketitle
\vspace{-4mm}
\begin{abstract}
Single-pixel imaging is a novel imaging scheme that has gained popularity due to its huge computational gain and potential for a low-cost alternative to imaging beyond the visible spectrum. 
The traditional reconstruction methods struggle to produce a clear recovery when one limits the number of illumination patterns from a spatial light modulator. As a remedy, several deep-learning-based solutions have been proposed which lack good generalization ability due to the architectural setup and loss functions. In this paper, we propose a generative adversarial network based reconstruction framework for single-pixel imaging, referred to as SPI-GAN. 
Our method can reconstruct images with 17.92 dB PSNR and 0.487 SSIM, even if the sampling ratio drops to $5\%$. This facilitates much faster reconstruction making our method suitable for single-pixel video. 
Furthermore, our ResNet-like architecture for generator leads to useful representation learning that allows us to reconstruct completely unseen objects. The experimental results demonstrate that SPI-GAN achieves  significant performance gain, e.g. near 3dB PSNR gain, over the current state-of-the-art method.      
\end{abstract}
\section{Introduction}  
Single-pixel imaging (SPI) \cite{duarte2008single} has drawn wide attentions due to its effectiveness in cases where array sensors are expensive or even unavailable. The foundation of this imaging technique lies in the mathematics and algorithms of \emph{compressive sensing} \cite{donoho2006compressed}. SPI has been widely used in applications such as spectral imaging \cite{li2017efficient,wang2015hyperspectral,bian2016multispectral}, optical encryption \cite{clemente2010optical,chen2013ghost}, lidar-based sensing \cite{zhao2012ghost,gong2016three}, 3D imaging \cite{sun20133d,sun2016single}, imaging in atmospheric turbulence \cite{cheng2009ghost,zhang2010correlated}, object tracking \cite{magana2013compressive,li2014ghost,gibson2017real}, etc. SPI has a similar imaging technique like ghost imaging (GI) \cite{pittman1995optical,strekalov1995observation}. The GI technique has been shown to be applicable to classical thermal light \cite{ferri2005high,bennink2002two}. To reduce the computation time, \cite{shapiro2008computational} proposed to use a spatial light modulator (SLM) to modulate light; which coincides with the idea of the SPI. With the help of a SLM and a photon detector, one can employ the SPI technique by illuminating different patterns onto an image and later capturing the reflected light. A converter digitizes the detector's output, also known as measurements, and dispatches them to the reconstruction process.
In general, the performance of SPI depends on the number of illumination patterns used for encoding the information in the image. However, the acquisition time increases linearly with the number of illumination patterns. Several efforts have been made to come up with the solution for reducing the acquisition time. Recently, a fast random pattern illumination technique has been developed for faster acquisition of objects \cite{wang2017high, edgar2015simultaneous}. 

Efforts have been made to reconstruct target images that are acquired through SPI \cite{gong2015high, hu2015patch, yu2014adaptive}. There are several iterative \cite{tropp2007signal, donoho2006compressed, candes2006stable,yang2010fast} and non-iterative \cite{ferri2010differential, gong2010method, shin2016performance} approaches that are usually deployed for such reconstruction. One has to formulate the SPI as an error minimization problem for techniques described in \cite{5666245, liao2014generalized}. Whereas, methods like differential ghost imaging (DGI) \cite{ferri2010differential} exploit the correlation information between the illumination patterns and the measurements. However, the quality of the reconstruction is poor compared to iterative approaches. In addition, non-linear optimization are used to solve this problem in \cite{abetamann2013compressive,duarte2008single,suo2016signal}. These methods use prior knowledge about the image to solve the inverse problem in an iterative fashion. Due to the iterative process, they demand high computational cost and longer reconstruction time both of which are not desirable in practical scenarios. In addition, these methods achieve sub optimal performance even for a reasonable number of measurements. As a viable and better alternative to these methods, there has been a growing interest in incorporating \emph{deep learning for SPI}.


\begin{figure*}[htb]
\centering
   \includegraphics[width=0.85\linewidth]{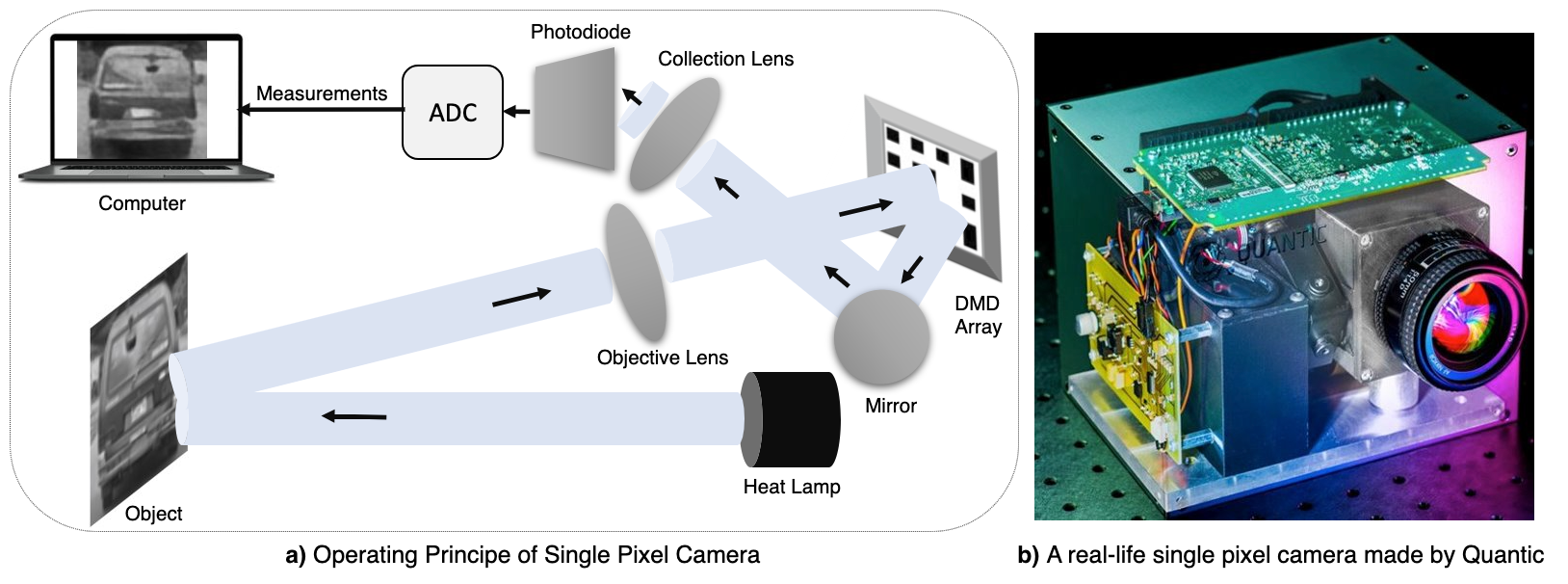}
   \caption{a) Image acquisition and reconstruction process using a single-pixel camera. The light from the scene comes through a objective lens to the digital micro-mirror device (DMD) array instead of multi million pixel sensors in a conventional camera.
   After converting the voltages in the photodiode to digital values, a suitable reconstruction method is used for object recovery. b) Photo Courtesy: Quantic \cite{quantic}. }
\label{fig:single_pixel_camera}
\end{figure*}
For example, GIDL \cite{lyu2017deep} proposes a fully connected network (FCN) to reconstruct an image from their corresponding CS measurements. However, FCN does not bid well in image-related tasks and fails to reconstruct natural images. Later,
DLGI \cite{wang2019learning} proposed a convolution neural network (CNN)  
that recovers images directly from the measurements that do not contain any spatial correlation information of the original image. This makes it harder to reconstruct test images that are not seen during training time.
Deepghost \cite{rizvi2020deepghost} proposes an autoencoder-based reconstruction scheme. However, the representation power of such an autoencoder is limited due to its shallow architecture. This results in poor generalization performance for test images as well as other unseen images.
In addition, the performance of these DL-based methods deteriorate rapidly in case of less number of measurements. Designing networks that offer to learn useful feature representation of the input data could be a remedy to aforementioned challenges. It should be noted that the structure and depth of a neural network mostly determines the representation power of it. Furthermore, the way one optimizes the network parameters can impact the reconstruction performance significantly. Most of the DL-based reconstruction methods employ mean squared error (MSE) loss to optimize the network. However, MSE tends to produce blurry output and fails to recover high frequency contents in the image. In this work, we aim to consider multiple loss functions as it has been shown in \cite{yang2019deep,ledig2017photo} that proper choice of loss functions can lead to high-quality image recovery. 


We propose a GAN-based reconstruction technique for single-pixel imaging, referred to as $\emph{SPI-GAN}$. Our aim is to reconstruct images through the generative modelling of the reconstruction process. In general, a GAN consists of two parts: a generator network that generates fake images and a discriminator network that learns to differentiate between the real and fake images. In our setup, the generator tries to \emph{reconstruct the original images} from the noisy inputs and the discriminator treats the \emph{reconstruction images as fake images}. Through minimizing a carefully designed loss function, our reconstruction framework harnesses the ability of learning useful and robust representations from noisy images to produce a clear reconstruction. Moreover, our findings indicate that having skip connection improves the training experience as well as the reconstruction performance. It also facilitates a deep architecture for the generator network which accommodates better representation learning. In contrast to other methods, SPI-GAN achieves significant performance gain in terms of both peak signal-to-noise ratio (PSNR) and structural similarity index (SSIM). Furthermore, our method shows robustness in the presence of high measurement noise. 
Our main contributions can be summarized as follows: 
\begin{itemize}
\item We design a novel DL-based reconstruction framework to tackle the problem of high-quality and fast image recovery in single-pixel imaging. We show that proper choice of architecture and objective functions can lead to a better reconstruction. \item Due to the fast image recovery, our method is also applicable to \emph{single-pixel video} applications. We demonstrate this by reconstructing videos from a large and diverse dataset at a high frame per second (fps). \item Unlike other DL-based reconstructions, SPI-GAN shows better generalization ability to completely unseen datasets; which can be immensely helpful in many practical scenarios. To prove this, we train our network on STL-$10$ dataset \cite{coates2011analysis} and test it on $6$ completely different unseen datasets.
\end{itemize}
The rest of this paper is organized as follows: Section \ref{sec:Related_work} reviews the related studies. Section \ref{sec: SPI_model} gives an overview on single-pixel camera and its mathematical modelling. The SPI-GAN framework is introduced in Section \ref{sec:framework}. In Section \ref{sec:experiments}, we present the experimental results and conclude the paper in Section \ref{sec:conclusion}. 
\section{Related Work} \label{sec:Related_work}
In this section, we briefly review the iterative and deep-learning-based solutions for SPI followed by an overview on the applications of GAN. 

\subsection{Reconstruction Methods}
Conjugate gradient descent (CGD) \cite{5666245} is an iterative algorithm that treats the image reconstruction as a quadratic minimization problem. The minimization is carried out using the gradient descent approach. The goal is to minimize the loss function by iteratively updating the reconstruction image using loss gradient. In CGD, these gradients are designed to be conjugate to each other. Because of this, the conjugate gradient descent converges faster than the vanilla gradient descent method. The alternating projection (AP) \cite{liao2014generalized} is based on the theory of spatial spectrum. Each of the measurements can be considered as the spatial-frequency coefficient of the arriving light field at the photodiode. With the help of support constraints for Fourier and spatial domains, the reconstruction process alternates between these domains to update the recovered image.  Sparse-based methods \cite{duarte2008single} formulates the single pixel imaging problem as an $\ell_1$  minimization problem. Using linear programming based approaches such as basis pursuit \cite{4016283}, we can solve this convex optimization problem.

In different imaging techniques, deep learning methods are deployed to solve the inverse problem of recovering images. For example, DL is being employed in digital holography \cite{ren2018learning,wang2018eholonet,rivenson2018phase}, scattering media based imaging \cite{lyu2019learning,li2018deep,li2018imaging}, lenseless imaging \cite{sinha2017lensless}, and fluorescence lifetime imaging \cite{wu2016artificial}. In addition to this, recent DL-based methods have also shown great success in recovering images in an SPI setup~\cite{lyu2017deep, he2018ghost, wang2019learning, rizvi2020deepghost}. In most cases, these methods are better and faster compared to the conventional iterative algorithms. For example, GIDL \cite{lyu2017deep} employs a fully connected network (FCN) for reconstruction.  Later, a more appropriate convolution-based neural network (CNN) based approach was introduced in ~\cite{wang2019learning}, referred to as DLGI, which can successfully reconstruct simple images such as handwritten digits.  Another deep-learning-based approach, called Deepghost \cite{rizvi2020deepghost}, proposed an autoencoder-based  architecture to recover images from their undersampled reconstructions. Deepghost is based on the principle of denoising operation and shows promising results. 

\subsection{Applications of GAN}
Over the years, GAN \cite{goodfellow2014generative} has been used in many computer vision related tasks\cite{zhu2017unpaired, isola2017image, mirza2014conditional} as well as in the natural language domain \cite{rajeswar2017adversarial,press2017language}. Furthermore, GAN has also been used for reconstructing medical images\cite{seeliger2018generative, shen2019deep, shen2019end}, face images \cite{vanrullen2019reconstructing} etc. GAN has proved to be an effective method to produce high-quality images compared to other neural networks. In recent times, GAN has attracted a lot of attentions because of its effectiveness in producing high-quality images \cite{ledig2017photo, isola2017image}. In addition, there have been many scenarios \cite{reed2016generative, denton2015deep, zhu2017unpaired} where GAN is able to learn complex data distributions. Unlike normal generative models, which face difficulties in approximating intractable probabilistic computations, GAN trains generative models that are better suited in approximating data distribution. In general, the generator tries to fool the discriminator by producing samples close to real data. On the other hand, the discriminator network tries to differentiate between the ground truth and the generated output. The training of a GAN resembles that of a min-max game where the generator learns a data distribution that is close to the real data distribution. These are few among many factors that motivate us in designing our reconstruction framework based on GAN.

\section{Single-Pixel Imaging (SPI)}\label{sec: SPI}
Fig. \ref{fig:single_pixel_camera} shows the setup of a single-pixel camera. It consists of two main components: the single-pixel detector and the spatial light modulator (SLM). There are different technologies available that can be employed for the modulation technique. One such technology is digital light projectors (DLPs) \cite{sampsell1994digital} which are usually based on the digital micro-mirror device (DMD). The DMD arrays that are usually employed in single-pixel imaging are known for their superior modulation rates, 20 kHz. In general, there are $1024 \times 768$ number of mirrors present in a DMD array that can be spatially oriented for each pixel of an image.   

In addition to these components, imaging through single-pixel camera requires additional optical lenses, guiding mirror, analog-to-digital (A/D) converter etc. Let us Consider that we want to obtain the image of a scene or an object via a single-pixel camera. First, we illuminate the object with the help of a light source. An objective lens focuses the reflected light from the scene towards the DMD. In general, the DMD consists of an array of mirrors that operates as the scanning basis or the measurement matrix. For each pixel in the object, a mirror or a subset of mirrors are oriented towards or away from the reflected light. This operation is equivalent to multiplying the image with binary masks (0 or 1)~\cite{duarte2008single}. We put 0, at the scanning basis, for mirrors that are oriented away from the lens and 1 for the mirrors with orientation towards lens. The orientation of the mirrors causes some areas of the image to be masked while reflecting the light in other areas. A beam steering mirror focuses the reflected light to a photodiode (the single detector or sensor) through the collection lens. The detector collects this light as a form of voltages that are digitized using an A/D converter to numeric bitstream. We repeat above steps for $K$ number of times to encode enough information about the object. However, a different binary pattern needs to be generated at each time to comprise a total of $K$ measurements. Finally, an appropriate reconstruction algorithm is employed for reconstructing the scene from measurements. The focus of our work is the accurate reconstruction of a scene with as few measurements as possible.

\subsection{SPI Model} \label{sec: SPI_model}
In this section, we present the mathematical modelling of a single-pixel camera. The foundation of SPI technique lies in the domain of compressed sensing (CS) \cite{candes2006stable}. In compressed sensing, it has been proven that a compressible signal or image $\boldsymbol{x} \in \mathbb{R}^{W \times H}$ is recoverable from relatively fewer random projections, $\boldsymbol{y} \in \mathbb{R}^{K}$. Where $W$ and $H$ are the dimensions of $\boldsymbol{x}$ and $K$ is the number of random projections. As convention, we interchangeably use notation $W \times H$ with $N$ as both stands for the total number of pixels in $\boldsymbol{x}$. 

In refer to the single-pixel camera, $\boldsymbol{x}$ can be considered as the object in Fig. \ref{fig:single_pixel_camera} and $\boldsymbol{y}$ represents the measurements from the photodiode. To get $\boldsymbol{y}$, we need to generate $K$ number of $W \times H$ binary masks or patterns in the DMD array. In general, total number of patterns ($K$) is much smaller than the total number of pixels ($N$); that is $K/N << 1$. The inner product of these binary masks and $\boldsymbol{x}$ gives us the measurements, 
\begin{equation}
    \boldsymbol{y} = \boldsymbol{\Phi} \boldsymbol{x} + \boldsymbol{q}. 
\end{equation}
Here, $\boldsymbol{\Phi} \in \mathbb{R}^{K \times N}$ is the normalized binary scanning basis for SPI and $\boldsymbol{q}$ accounts for measurement noise. The noise term $\boldsymbol{q} \in \mathbb{R}^{K} $ can be modelled as additive white Gaussian noise (AWGN) with $\boldsymbol{q} \sim \mathcal{N}(\boldsymbol{0},\sigma^2 \boldsymbol{I}_K)$, where $\sigma$ stands for the standard deviation and $\boldsymbol{I}_K$ is an all-one vector. We describe more about measurement noise in the experiment section. Now, consider a scenario where we can represent $\boldsymbol{x}$ in terms of an orthonormal basis $\boldsymbol{\psi} \in \mathbb{R}^{N}$ as,
\begin{equation}
    \boldsymbol{x} = \sum_{i=1}^{N} \boldsymbol{s_{i} \psi_{i}} = \boldsymbol{\Psi} \boldsymbol{s}
\end{equation}
, where $\{\boldsymbol{\psi_1, \dots, \psi_N} \}$ are the column vectors of an $N \times N$ basis matrix $\boldsymbol{\Psi}$. The updated equation for $\boldsymbol{y}$ is given by
\begin{equation}
    \boldsymbol{y} = \boldsymbol{\Phi} \boldsymbol{x} + \boldsymbol{q}
    = \boldsymbol{\Phi} \boldsymbol{\Psi} \boldsymbol{s} + \boldsymbol{q} =  \boldsymbol{\Theta} \boldsymbol{s} + \boldsymbol{q}.
\end{equation}  
One can use both iterative and DL-based reconstruction algorithms to recover $\hat{\boldsymbol{x}}$ from $\boldsymbol{y}$. Compared to DL-based methods, iterative methods such as conjugate gradient descent (CGD) \cite{5666245} or alternating projection (AP) \cite{liao2014generalized} require higher number of measurements to produce a good reconstruction. Furthermore, they need longer reconstruction time in contrast to DL-based methods. In general, object $\boldsymbol{x}$ can lie in different domains \cite{zhao2012ghost,chen2013ghost} depending on the applications of SPI. In our work, we present the model for natural images that can be generalized to images from other domains too. 


\begin{figure*}
    \centering
    \includegraphics[width=1\linewidth]{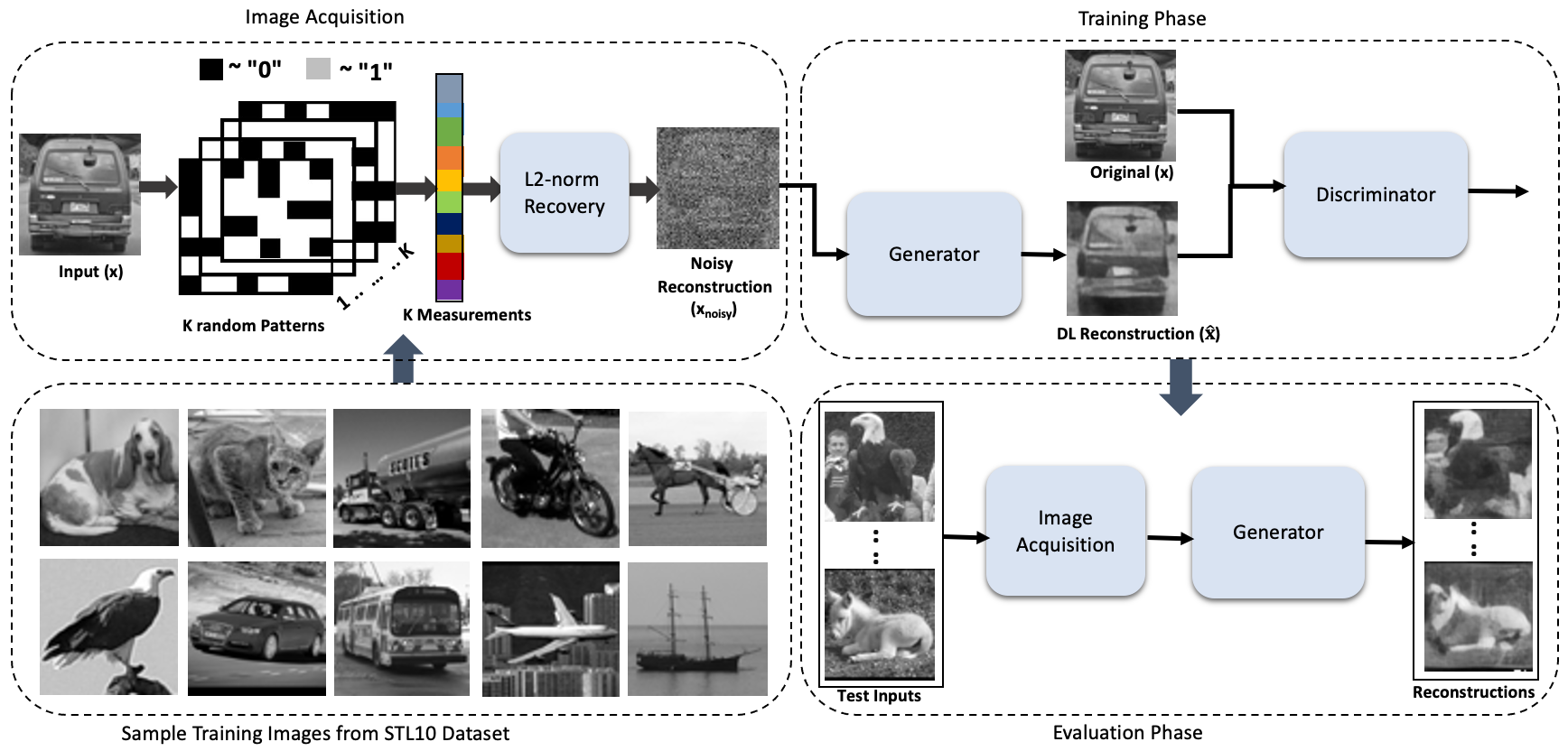}
    \caption{Our proposed SPI-GAN framework mainly consists of a generator that takes the noisy $\ell$2-norm solution ($\hat{\boldsymbol{x}}_{noisy}$) and produce a clear reconstruction ($\hat{\boldsymbol{x}}$) that is comparable to $\boldsymbol{x}$. On the other hand, a discriminator learns to differentiate between $\boldsymbol{x}$ and $\hat{\boldsymbol{x}}$ in an attempt to not to be fooled by the generator.}
    \label{fig:summary_figure}
\end{figure*}

\subsection{Deep-learning-based reconstruction} \label{sec:DL_recons}
For tasks such as image reconstruction, it is customary to employ convolution neural network (CNN) based architectures. The primary operation that we rely on is convolution between the image and a set of filters. This resembles the process of extracting patches from images and representing them by a set of basis. These bases can be produced by principal component analysis (PCA) \cite{wold1987principal}, discrete cosine transform (DCT) \cite{ahmed1974discrete}, etc. In our setup, we define these bases as convolution filters that need to be optimized. In a CNN-based architecture, there can be several convolution layers with activation, pooling, and batch normalization in each layer. At the end of the network, there is a loss function to measure the difference between input and output. In general, a pixel-wise mean square error (MSE) loss function is used for tasks like reconstruction. MSE loss is optimized using a gradient-based optimizer by back-propagating the loss gradient through the network. After training, we have fully optimized network parameters at hand. These parameters can be used readily for reconstructing images. Any DL-based methods usually follow these steps to train and optimize a network. However, the only loss function that these methods optimize is the MSE loss. While MSE loss has its advantages, the reconstructed targets tend to be blurry due to the average of possible solutions the network produces. To produce a more clear reconstruction, we can consider loss functions based on the perceptual similarity between $\boldsymbol{x}$ and $\hat{\boldsymbol{x}}$. Hence, we bring in the perceptual similarity loss that considers distance between features extracted by a pre-trained network; instead of distance in the pixel space. 


    

\section{SPI-GAN Framework} \label{sec:framework}
In this section, we present the details of the SPI-GAN framework. Our proposed technique consists of an image acquisition process followed by the training and evaluation of the GAN. 

Fig. \ref{fig:summary_figure} shows the setup of the SPI-GAN framework. First, we collect the training samples from a public dataset that contains a large number of natural images. For each image ${\boldsymbol{x}}$, we then generate $K$ random binary patterns $\boldsymbol{\Phi}$ which consists of $K$ number of rows drawn from a $0/1$ Walsh matrix. We then randomly permute these rows before normalizing them. After that, we perform a simple and noisy reconstruction employing the \emph{$\ell$2-norm recovery} method described in \cite{baraniuk2007compressive}. We can find the minimum $\ell$2-norm solution by solving
\begin{equation} \label{eq:l2_optimization}
\hat{\boldsymbol{s}} = \argmin ||\boldsymbol{s}'||_{2} \hspace{0.2cm} \text{such that} \hspace{0.2cm} \boldsymbol{\Theta s'}= \boldsymbol{y}.
\end{equation}
Solution of this optimization problem can be written as 
\begin{equation} \label{eq:l2_norm_solution}
    \hat{\boldsymbol{s}} = (\boldsymbol{\Theta}^{T} \boldsymbol{\Theta})^{-1}\boldsymbol{\Theta}^{T}\boldsymbol{y}.
\end{equation}
And this solution gives us the noisy $\ell$2-norm reconstruction, 
\begin{equation}
    \hat{\boldsymbol{x}}_{noisy} = \Psi \hat{\boldsymbol{s}}.
\end{equation}
This non-iterative method with a closed-form solution can recover images much faster than iterative-based approaches; even though the image quality is compromised due to the non-sparse solution, $\hat{\boldsymbol{s}}$.

However, we significantly enhance the quality of $\hat{\boldsymbol{x}}_{noisy}$ by feeding it to the \emph{generator network}, $G(.)$, that gives us our final reconstruction, $\hat{\boldsymbol{x}}$. 
The generator output $\hat{\boldsymbol{x}}$ and ${\boldsymbol{x}}$ are then fed to the discriminator, $D(.)$, that learns to differentiate between these two. Here, our main goal is to train both networks in an adversarial manner. After training, we disregard the discriminator and evaluate the generator as our main reconstruction network. Images in the evaluation phase are completely different from the training images. One of the essential parts of our framework is the design of the generator and discriminator networks that are shown in Fig. \ref{fig:generator_disc}. We describe more about them in the supplementary material. 

 \begin{figure*}[htb]
    \centering
    \begin{minipage}{0.8\textwidth}
    \includegraphics[width=1\linewidth]{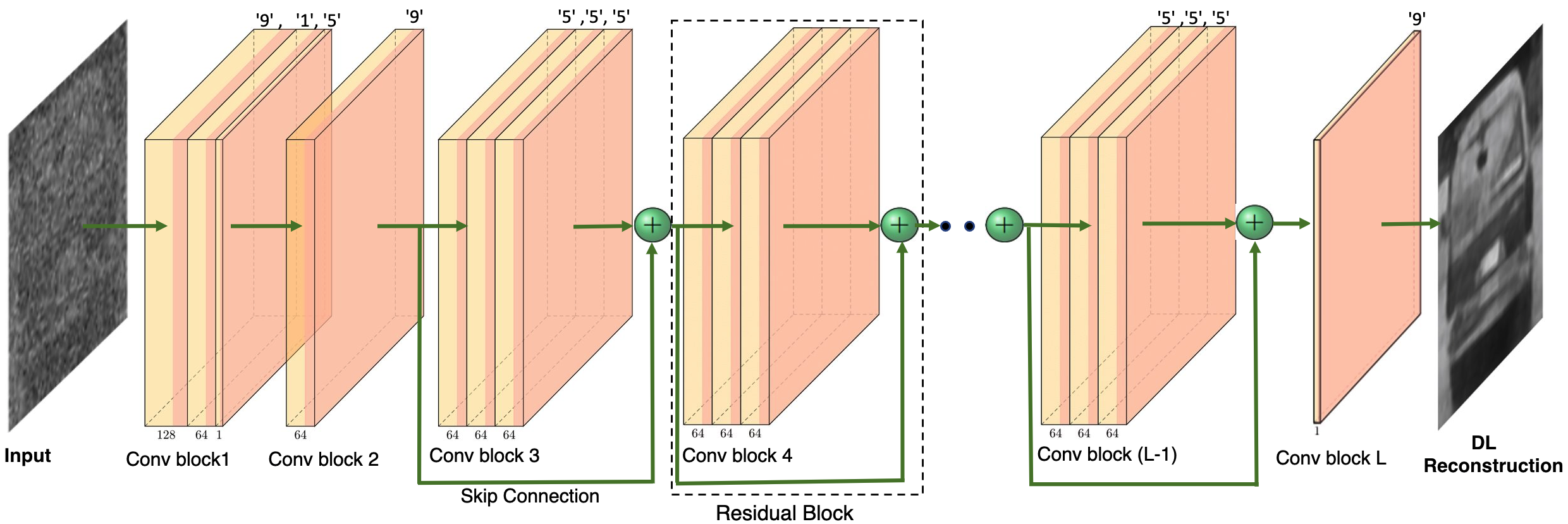}
    
    \end{minipage}
    \begin{minipage}{0.8\textwidth}
    \centering
    \includegraphics[width=1\linewidth]{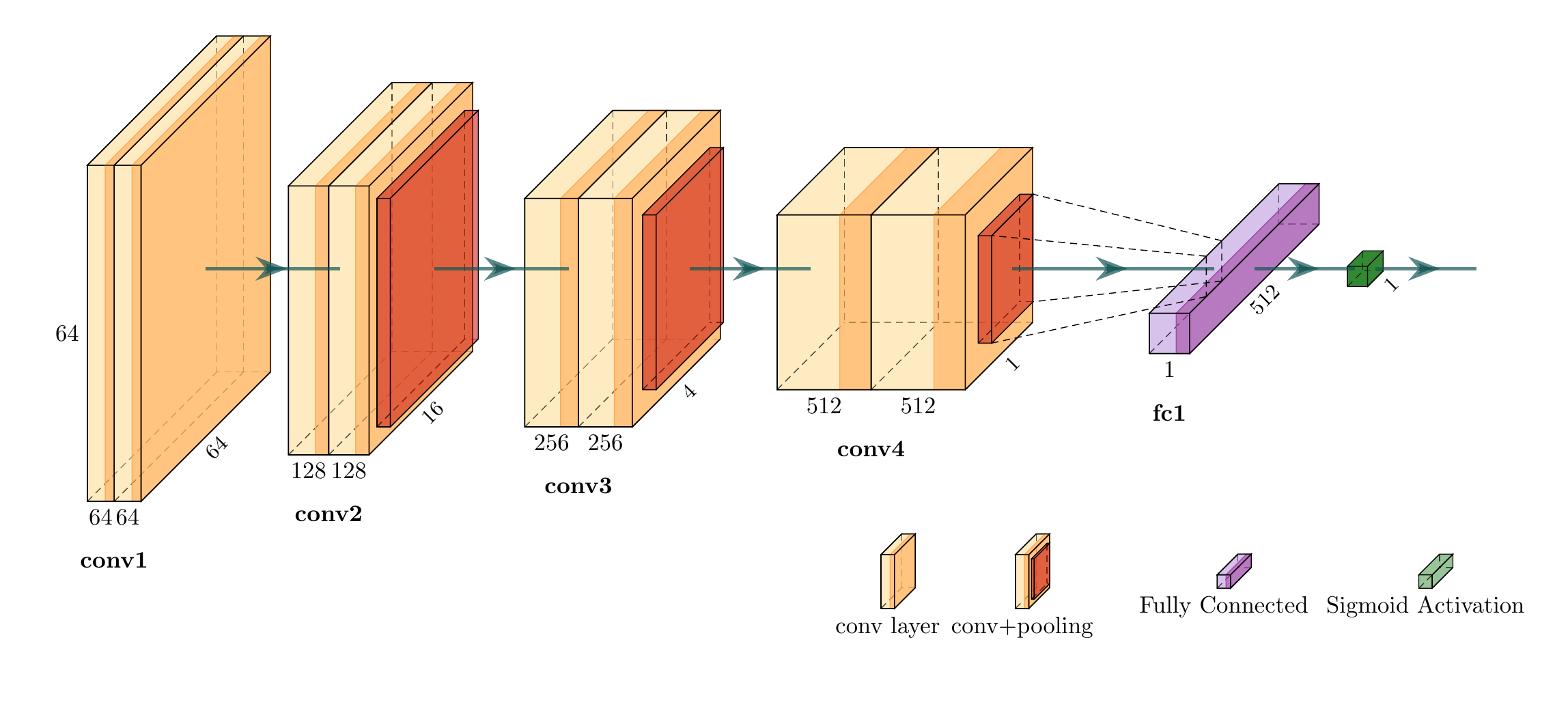}
    \end{minipage}%
    \vspace{-5mm}
    \caption{\emph{Top:} The generator network employs a ResNet-like architecture to produce the desired reconstruction. We use convolution layers with different kernel sizes and parametric ReLU activation to extract important features from the input. \emph{Bottom:} Discriminator network uses a VGG-like architecture \cite{simonyan2014very} that gradually increases the number of filters with pooling layers. Last convolution output goes to the fully connected layer followed by the sigmoid activation. At the end, a binary cross entropy loss function is optimized to discriminate between original image and the reconstructed image.}
    \label{fig:generator_disc}
\end{figure*}
\subsection{GAN Training}
For training, we first initialize the parameters of the two networks, $\Theta_{G}$ and $\Theta_{D}$. Here, $\Theta_{G}$ represents the weights and biases of the the convolution layers we stacked to construct the generator. Similarly, $\Theta_{D}$ bears the same notion for the discriminator network. The output of the generator then goes through the discriminator network whose job is to differentiate between the original image and the reconstructed image. The better the reconstruction of the generator, the higher the chance of the discriminator being fooled. To achieve this, this can be formulated as a min-max game where the discriminator tries to maximize its discrimination ability between $\boldsymbol{x}$ and $\hat{\boldsymbol{x}}$. On the other hand, the generator minimizes this discrimination power by fooling the discriminator to misclassify $\hat{\boldsymbol{x}}$ as $\boldsymbol{x}$. Following \cite{goodfellow2014generative}, the overall optimization problem is given by:   
\begin{equation} \label{eq:minmax}
     \min_{\Theta_G} \max_{\Theta_D} \mathrm{E}_{\boldsymbol{x}}[log D(\boldsymbol{x})] +
     \mathrm{E}_{\hat{\boldsymbol{x}}_{noisy}}[log (1- D(G(\hat{\boldsymbol{x}}_{noisy}))].
\end{equation}
Here, $\mathrm{E}$ stands for the expectation and $D(\boldsymbol{x})$ and $D(G(\hat{\boldsymbol{x}}_{noisy}))$ are discriminator outputs corresponding to the original and reconstructed image, respectively. The main idea is to jointly train $D(.)$ and $G(.)$ so that the generator can produce images, $\hat{\boldsymbol{x}}$, that are close to the real ones, $\boldsymbol{x}$. On the other hand, the discriminator learns to distinguish between $\hat{\boldsymbol{x}}$ and $\boldsymbol{x}$. This turns out to be a zero-sum game where one network functions as the adversary of the other network.  

Let, $\{\boldsymbol{x}^{(1)},....., \boldsymbol{x}^{(M)}\}$ is a minibatch of M samples taken from data distribution $p_{data}(\boldsymbol{x})$. We update the $\Theta_D$ through ascending its stochastic gradient,
\begin{equation} \label{eq:dis_params}
       \nabla_{\theta_D} \frac{1}{M} \sum_{m=1}^M \left[log D(\boldsymbol{x}) + log (1- D(G(\hat{\boldsymbol{x}}_{noisy}))\right].
\end{equation}
Similarly, we optimize for $\Theta_{G}$ through minimizing a loss over $M$ number of training samples $\{\hat{\boldsymbol{x}}_{noisy}^{(1)},....., \hat{\boldsymbol{x}}_{noisy}^{(M)}\}$. Thus, the gradient descent update can be expressed as,  
\begin{equation} \label{eq:gen_params}
   \nabla_{\theta_G} \frac{1}{M} \sum_{m=1}^M l^{rec}(G(\hat{\boldsymbol{x}}^{(m)}_{noisy}), \boldsymbol{x}^{(m)}),
\end{equation}
where $l^{rec}$ is the total reconstruction loss, which is a combination of three different loss functions. The first loss function is the \emph{pixel-wise MSE loss} that is defined as
\begin{equation} \label{eq:MSE_loss}
     l^{rec}_{MSE} = \frac{1}{WH} \sum_{i=1}^W \sum_{j=1}^H (\boldsymbol{x}_{i,j} - G(\hat{\boldsymbol{x}}_{noisy})_{i,j})^2.
\end{equation}
Minimizing this loss alone often results in good performance. However, pixel-wise MSE takes the average over possible solutions that results in blurry output. Furthermore, reconstructed output misses on the high frequency content making it perceptually unsatisfying \cite{mathieu2015deep, johnson2016perceptual}. As a forwarding step in solving this problem, we use a loss function that operates on the concept of perceptual similarity. We take a pre-trained VGG19 network and use it as the feature extractor for the $\hat{\boldsymbol{x}}_{noisy}$ and ${\boldsymbol{x}}$. Let, $Q_{k}(.)$ is the $k^{th}$ convolution-layer output and $W_k$ and $H_k$ represents the dimensions of the feature maps of the VGG network. The \emph{perceptual similarity loss} can be calculated for the $k^{th}$ layer as, 
\begin{equation}\label{eq:Q_loss}
     l^{rec}_{sim} (k) = \frac{1}{W_k H_k} \sum_{i=1}^{W_k} \sum_{j=1}^{H_k} (Q_k(\boldsymbol{x}_{i,j}) - Q_k(G(\hat{\boldsymbol{x}}_{noisy})_{i,j}))^2.
\end{equation}
We get the output feature maps $Q_k(G(\hat{\boldsymbol{x}}_{noisy}))$ and $Q_k(\boldsymbol{x})$  corresponding to the inputs $G(\hat{\boldsymbol{x}}_{noisy})$ and $\hat{\boldsymbol{x}}$, respectively. Depending on $k$, we take the activation output with or without max-pooling as some of the layers in VGG19 does not have max-pooling. 


Along with the MSE and the perceptual similarity loss, we define the \emph{adversarial loss} as a requirement for the training of GAN. The purpose of the adversarial loss, $l^{rec}_{adv}$ is to encourage solutions that are closer to the real data. As this loss is related to the discriminator network, which the generator apparently tries to fool, we minimize the negative log-probabilities of the discriminator for all training samples. Thus, the adversarial loss can be written as 
\begin{equation}\label{eq:adv_loss}
l^{rec}_{adv} = \frac{1}{M} \sum_{i=1}^M -log D(G(\hat{\boldsymbol{x}}^{(i)}_{noisy}))
\end{equation}
, where $G(\hat{\boldsymbol{x}}_{noisy})$ is the output of the generator and $D(G(\hat{\boldsymbol{x}}_{noisy}))$ is the probability for $\hat{\boldsymbol{x}}_{noisy}$ to belong in the manifold of the natural images. Minimizing this loss function is equivalent to maximizing the chance of fooling the discriminator. 

Finally, the loss function for optimizing $\Theta_{G}$ can be written as the weighted sum of three loss functions \cite{ledig2017photo},
\begin{equation}\label{eq:rec_loss}
l^{rec} = l^{rec}_{MSE} + \lambda_{sim} \times l^{rec}_{sim}(k)  + \lambda_{adv} \times l^{rec}_{adv}, 
\end{equation}
where we choose $k=11$. The reason we choose this large value of $k$ is because it facilitates the comparison of $\boldsymbol{x}$ and $\hat{\boldsymbol{x}}$ in the deep feature space. This in turn helps us to obtain better reconstruction performance. We scale the similarity and adversarial loss by $\lambda_{sim}$ and $\lambda_{adv}$, respectively. In our work, the loss coefficients  $\lambda_{sim}$ and $\lambda_{adv}$ are set to have values of $6e-3$ and $1e-3$ respectively. MSE loss has the highest importance as it is our primary loss function for reconstruction.

\begin{figure}
    \centering
    \includegraphics[width=0.8\linewidth]{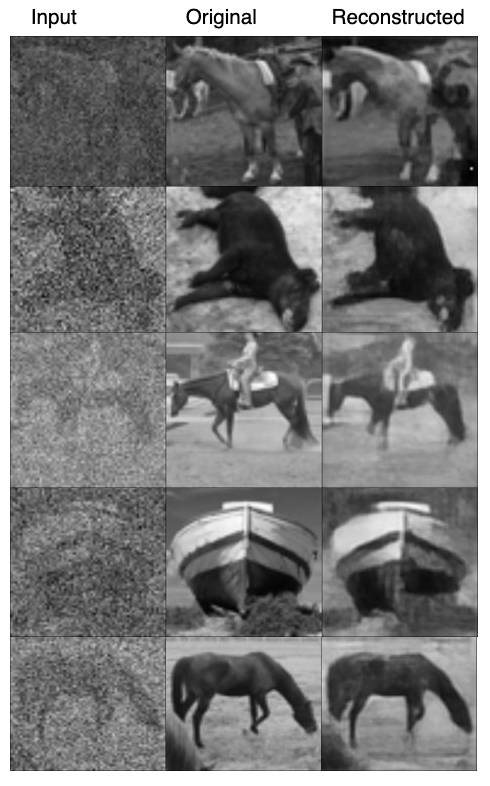}
    \caption{A comparison of images and their reconstruction. The input ($\hat{\boldsymbol{x}}_{noisy}$) is fed to the generator that gives us  the reconstructed ($\hat{\boldsymbol{x}}$) image. Sampling rate is fixed at 20 $\%$.}
    \label{fig:ratio_0.20}
\end{figure}

\begin{figure*}
    \centering
    \includegraphics[width=1\linewidth]{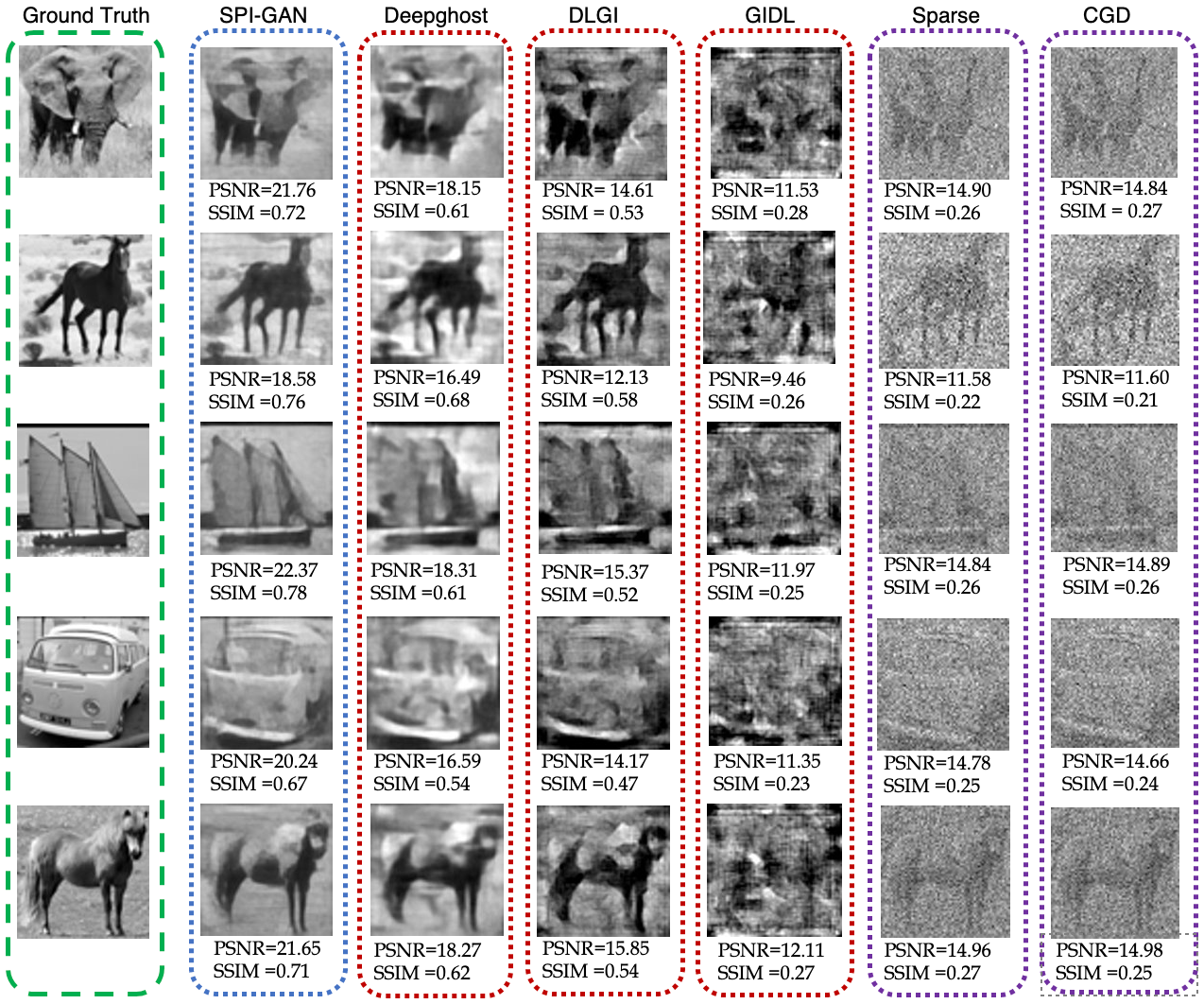}
    \caption{Reconstructed samples from the test dataset. SPI-GAN outperforms other DL-based methods in terms of both PSNR and SSIM. Sampling rate is set at 20 $\%$.}
    \label{fig:main_comparison}
\end{figure*}


\section{Experimental Results}\label{sec:experiments}
\subsection{Training Data Preparation}
As shown in Fig. \ref{fig:summary_figure}, training of the generator and discriminator requires a large number of images. These images can be collected by different means depending on the nature of applications \cite{zhao2012ghost,gong2016three,li2017efficient,wang2015hyperspectral,bian2016multispectral}. However, collecting thousands of images from these domains poses challenges like equipment limitations, rare imaging environment, etc. Instead, a solution for this can be found in the large image datasets such as ImageNet and STL10, that are publicly available for deep learning applications. Furthermore, natural images (animals, vehicles, etc.) collected using digital camera closely represent the images that are suitable for single-pixel imaging. This suggests that the performance obtained on these datasets should generalize to images from other domain too. Taking above factors into account, we only consider natural images in this work. 
\subsubsection{Image Dataset} 
We collected the necessary training and validation data from STL10 image dataset. STL10 contains 100,000 unlabeled images and 13,000 annotated images. All of these images are in RGB format. However, we convert them to single channel gray-scale image as we consider the camera model with a single guiding mirror. For RGB imaging, it is required to use multiple mirrors and color filters. The images in the STL10 dataset are of $96 \times 96$ size that are later resized to $64 \times 64$. Therefore, each image contains 4096 pixels. Some of the samples from STL10  are also shown in the Fig. \ref{fig:summary_figure}. For our experiment, we only chose 45,000 unlabeled images and split them into training, validation and test set. Among them, 40,000 images are used for training, 3,000 for validation and the rest of the images are test images. We then simulated the single-pixel camera where we used normalized binary patterns to encode an image into K number of measurements. The noisy $\ell_2$-norm solution is then fed to the generator. 
After training, we tested our network on images from validation and test sets that were unseen during training. We repeated this whole process for different sampling rates. The sampling rate (SR) can be defined as the ratio of the number of measurements ($K$) to the total number of pixels in the image. In our case, SR=$K/4096$.  
\subsubsection{Video Dataset} 
Apart from imaging, our method is also applicable for \emph{single-pixel video}. We use UCF101 action recognition dataset as the benchmark. With 13,320 videos from 101 action categories, UCF101 is one of the largest video datasets available. Since single-pixel camera can handle only one image per frame, we first extract the frames from the video and then feed them as images. These frames are then resized to $64 \times 64$ before we acquire and reconstruct them using SPI-GAN. We take 500 video clips as the training set and 50 videos as the validation set. 
\begin{figure}[htb]
    \centering
    \includegraphics[width=1\linewidth]{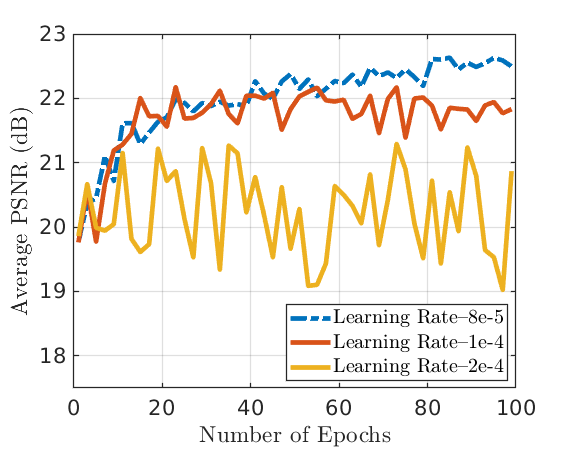}
    \caption{Average PSNR (dB) of the validation images  over the first 100 epochs of training. For stable training, the range of optimum learning rate is around $8e^{-5}$. Learning rate more or less than this may result in poor performance.}
    \label{fig:learning_final}
\end{figure}

\subsection{Performance Metric}
For evaluating the performance of our method, we employ two widely used metrics. The task at our hand requires to measure the image quality which is subjective and can vary from person to person. It is necessary to establish quantitative/empirical measures for image reconstruction algorithm like ours. Therefore, we choose peak signal-to-noise ratio (PSNR) and structural similarity index (SSIM) \cite{wang2004image} as the performance metrics. The higher the PSNR and SSIM values, the closer the reconstructed images are to the originals. The term  PSNR stands for the ratio between the peak amplitude of a clean image and the power of the distortion affecting the quality of that clean image. Because of the wide dynamic range of a signal or image, it is usual to take the logarithmic of PSNR in decibel (dB). Considering $\boldsymbol{x}_{ij}$ as the pixel value at $i^{th}$ row and $j^{th}$ column in image $\boldsymbol{x}$, the expression for PSNR is given by 
\begin{equation}
    PSNR = 10 log10(\frac{\boldsymbol{x}^{max}}{MSE}),
\end{equation}
where $\boldsymbol{x}^{max} = max_{i,j}\boldsymbol{x}_{ij}$. MSE is the mean square error between $\boldsymbol{x}^{max}$ and the reconstructed image $\hat{\boldsymbol{x}}$ and can be calculated as 
\begin{equation}
    MSE = \frac{1}{CWH}\sum_{i=1}^{W}\sum_{j=1}^{H}(\boldsymbol{x_{ij}}- \hat{\boldsymbol{x}}_{ij})^2,
\end{equation}
where $C$, $W$, and $H$ represent the number of channels, width, and height of an image, respectively. This term gives us the per-pixel loss of a reconstructed image.  

In addition to PSNR, SSIM is a good estimator for assessing the image quality. The range of SSIM lies in [-1,+1], where +1 indicates that two images are very similar and -1 is if they are very different. The ultimate goal is to reconstruct a good quality image with high PSNR and SSIM. We follow \cite{1284395} to calculate SSIM for comparing images based on their luminance, contrast, and structure. Mean and standard deviation of the 2 images are two important parameters to consider for these estimations. 
\begin{table*}
    \caption{Average PSNR (dB) of the 2000 test images for different reconstruction methods. The performance improves as we take more number of measurements. Top 3 rows shows the performance of iterative based approaches. Both iterative and DL-based methods perform worse than our method.}
  \centering
  \begin{tabular}{c|c|c|c|c|c|c}
     Sampling Rate & 5$\%$ & 10$\%$ & 15$\%$ & 20$\%$ & 25$\%$ & 30$\%$ \\
    \hline
     CGD \cite{5666245} & 13.20$\pm$2.583 & 13.51$\pm$2.485 & 14.27$\pm$2.496 &  14.75$\pm$2.447 & 14.75$\pm$2.628 & 15.05$\pm$2.531  \\ 
     Sparse \cite{duarte2008single}  & 13.13$\pm$2.592 & 13.67$\pm$2.812 &  14.11$\pm$2.296 &  14.48$\pm$2.401 & 14.76$\pm$2.528 & 15.16$\pm$2.783  \\ 
     AP \cite{liao2014generalized} & 13.10$\pm$2.813 & 13.42$\pm$2.352 &  14.02$\pm$2.624 &  14.33$\pm$2.534 & 14.61$\pm$2.331 & 14.90$\pm$2.463  \\ 
     GIDL \cite{he2018ghost} & 9.674$\pm$1.258 & 10.14$\pm$1.683 &  10.97$\pm$1.856 &  11.60$\pm$2.079 & 12.07$\pm$2.109 & 12.30$\pm$2.322\\ 
     DLGI \cite{wang2019learning}& 11.93$\pm$1.211 &13.22 $\pm$1.728 &  14.46$\pm$1.875 &  14.78$\pm$2.170 & 15.24$\pm$2.284 & 15.62$\pm$2.446  \\ 
     Deepghost \cite{rizvi2020deepghost}  & 14.91$\pm$1.458 & 16.43$\pm$1.403 &  17.35$\pm$1.752 &  17.74$\pm$1.607 & 18.14$\pm$1.679 & 18.62$\pm$1.734 \\ 
     SPI-GAN  & \textbf{17.92$\pm$1.792} &  \textbf{18.41$\pm$1.741}   &  \textbf{20.32$\pm$1.638} &  \textbf{21.11$\pm$1.925} & \textbf{21.42$\pm$1.987} & \textbf{21.87$\pm$2.153} \\
    \hline
  \end{tabular}

    \label{tbl:PSNR}
\end{table*}
\begin{table*}
    \caption{Average Structural Similarity Index (SSIM) of the 2000 test images for different reconstruction methods. Bottom 4 rows represents the performance of DL-based methods.}
  \centering
  \begin{tabular}{c|c|c|c|c|c|c}
     Sampling Rate & 5$\%$ & 10$\%$ & 15$\%$ & 20$\%$ & 25$\%$ & 30$\%$ \\
    \hline
     CGD \cite{5666245} & 0.154$\pm$0.037 & 0.175$\pm$0.051 &  0.195$\pm$0.063 &  0.217$\pm$0.070 & 0.249$\pm$0.064 & 0.261$\pm$0.071 \\ 
     Sparse \cite{duarte2008single} &  0.163$\pm$0.053 & 0.178$\pm$0.067 &  0.198$\pm$0.055 &  0.218$\pm$0.061 & 0.243$\pm$0.072 & 0.267$\pm$0.071  \\ 
     AP \cite{liao2014generalized} &  0.140$\pm$0.039 & 0.159$\pm$0.042 &  0.180$\pm$0.046 &  0.203$\pm$0.053 & 0.226$\pm$0.059 & 0.246$\pm$0.065  \\ 
     GIDL \cite{he2018ghost} &  0.122$\pm$0.021 & 0.133$\pm$0.027 &  0.141$\pm$0.035 &  0.169$\pm$0.019 & 0.186$\pm$0.014 & 0.213$\pm$0.027  \\ 
     DLGI \cite{wang2019learning} &  0.236$\pm$0.030 &  0.323$\pm$0.046 & 0.395$\pm$0.057 &   0.427$\pm$0.069 & 0.467$\pm$0.065 & 0.502$\pm$0.072 \\ 
     Deepghost \cite{rizvi2020deepghost} &  0.316$\pm$0.064 & 0.398$\pm$0.079 &  0.495$\pm$0.086 &  0.561$\pm$0.088 & 0.584$\pm$0.084 & 0.615$\pm$0.079 \\ 
     SPI-GAN  & \textbf{0.487$\pm$0.081} &  \textbf{0.534$\pm$0.085}   &  \textbf{0.627$\pm$0.069} &  \textbf{0.665$\pm$0.076} & \textbf{0.673$\pm$0.080} & \textbf{0.682$\pm$0.083} \\
    \hline
  \end{tabular}
    \label{tbl:SSIM}
\end{table*}
\subsection{Hyper-parameter Settings}
In this work, we use PyTorch as our machine learning framework. The training takes place on two NVIDIA 1080 Ti GTX GPUs. In each iteration, we choose 64 images per mini-batch and avoid using large batch sizes. The training period stretches for 150 epochs and an Adam optimizer with a learning rate of $8e{-5}$ was employed for training. A regulizer with a regularization coefficient of $5e^{-4}$ is used for better test-time performance. The learning rate plays an important role in training both generator and discriminator networks. Even though a high learning rate leads to faster convergence, it also can make the training process unstable. To choose the right range of learning rate, we have analyzed the effect of learning rate in the training experience. The performance for different learning rates is shown in Fig. \ref{fig:learning_final}. For a learning rate of $2e{-4}$, the training becomes unstable and leads to poor performance. As we decrease the learning rate the network learns in a steady manner. After validating several ranges of learning rate, we settled on a learning rate of $8e{-5}$ for the training. Upon the finalizing the hyper-parameters, we started our training about which we describe more in the supplementary section. Fig. \ref{fig:sampling_final} demonstrates the validation performance of our network as the training progresses. The results shown here is the first 100 epochs of training. The PSNR and SSIM curves follow the same rising pattern indicating steady learning process irrespective of the number of measurements. 
\subsection{Training Initialization}\label{sec:train_ini}
The generator and the discriminator networks consist of both convolution and fully connected layers. The generator network has 17 convolution blocks in which 14 of them are residual blocks. We use batch normalization in most of these layers which helps us in reducing the effect of internal covariate shift. This phenomena occurs during training due to the changes in the distribution of non-linear inputs. Generally, it is integrated in NN before every nonlinear activation and requires two parameters: one for the scaling and another is for the shifting. These two learnable parameters are also updated at each iteration. In the absence of batch normalization, training the network becomes more challenging. 

As for the loss functions, the pixel-wise MSE loss is very straightforward to calculate. The objective of MSE is to fill in the missing parts of the input image. On the other hand, the perceptual similarity loss, $l^{rec}_{sim}$, in (\ref{eq:Q_loss}), requires a pre-trained VGG19 network that is easily downloadable using PyTorch. Since we are only using this network for calculating the similarity loss, there is no need to re-train it. Using this network, we get the feature representations for both the reconstructed and the original image. The difference between their high dimensional representations helps the network to learn more about their perceptual similarity. For adversarial loss, $l^{rec}_{adv}$, we take the response of the discriminator to the generator's output. Finally, we jointly minimize all these 3 losses for optimized $\hat{\theta}_G$. On the other hand, We employ a simple cross entropy loss for optimizing $\Theta_{D}$, where we label  $\boldsymbol{x}$ as $'1'$ and $\hat{\boldsymbol{x}}$ as $'0'$ 
After feeding $\boldsymbol{x}$ to the discriminator, we compute the cross-entropy (CE)-loss based on the $D(\boldsymbol{x})$. Similarly, we calculate another CE-loss for $\hat{\boldsymbol{x}}$ using the output $D(\hat{\boldsymbol{x}})$. These two CE-losses are then jointly optimized to get the optimized $\hat{\Theta}_{D}$.

\begin{figure*}
    \centering
    \includegraphics[width=1\linewidth]{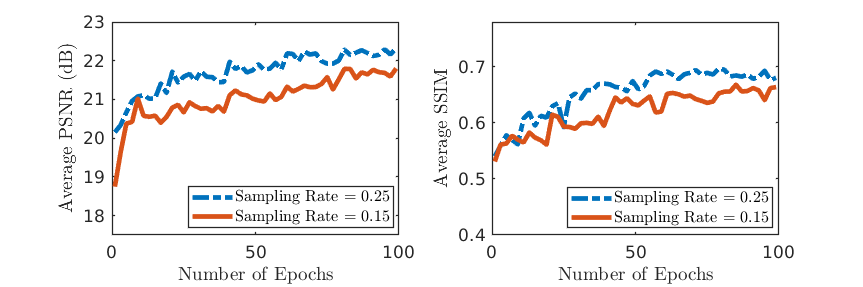}
    \caption{Average PSNR and SSIM for different sampling rates. As the training progresses, we get higher validation gains. Taking higher number of measurements facilitates better reconstruction.}
    \label{fig:sampling_final}
\end{figure*}

\subsection{Comparison with other methods}
In this section, we present the performance of SPI-GAN as well as other recovery methods. Using both quantitative (PSNR and SSIM) and qualitative measures, we show how our method outperforms other methods. Tables \ref{tbl:PSNR} and \ref{tbl:SSIM} show the performance comparison of all methods for different sampling rates. It can be observed that SPI-GAN outperforms all other methods for a wide range of sampling rate. We evaluate all these methods on a test set containing 2000 images. Our method obtains an average PSNR of 21.11 dB and an average SSIM of 0.665 for a sampling rate of $20\%$. These gains varies with the number of measurements. We decreased the sampling rate gradually to show the performance deterioration. Even for a sampling rate of $5\%$, we are able to recover images with PSNR close to 18 dB and 0.48 SSIM. This shows the effectiveness of SPI-GAN in applications where it may not be feasible to collect a lot of measurements. This also helps in faster reconstruction making our method suitable not only for images but also for video reconstructions. 
For a wide range of sampling rates, obtaining high PSNR and SSIM indicates that our method can reconstruct most of the test images quite satisfactorily. Even with very few measurements, the recovered images have high perceptual similarity to the ground truth. We discuss more on this in the supplementary material.  

The performance of CGD \cite{5666245} for different sampling rates are shown in Tables \ref{tbl:PSNR} and \ref{tbl:SSIM}. The average PSNR obtained for a 25$\%$ sampling rate is only 14.75 dB compared to 21.42 dB for our method. This performance gain is consistent for all sampling rates. Regarding SSIM, we obtain better performance for higher sampling rate. However, even for a 30$\%$ SR, average SSIM is only 0.261. 
Even though \cite{duarte2008single} employs rigorous optimization, it obtains only an average of 14.48 dB PSNR for a 20$\%$ sampling rate. The performance does not improve even for higher number of measurements. Same scenario can be observed for SSIM values. The final iterative method we considered here is the alternating projection (AP) \cite{liao2014generalized}. 
Like other two methods, AP also underperforms SPI-GAN in terms of both PSNR and SSIM. In terms of PSNR, AP performs on the same level as other iterative methods. For example, average PSNR values for all of these methods are around 15 dB for 30$\%$ sampling rate. However, in case of SSIM, it performs worse compared to other two iterative methods; averages only 0.203 and 0.246 for sampling rates of 20 $\%$ and 30 $\%$, respectively.   

\begin{table*}
    \caption{Average PSNR gain (dB) of different deep learning based methods under different noise levels.}
  \centering
  \begin{tabular}{c|c|c|c|c|c|c|c|c}
    
    Noise Level & 1e-4 & 3e-4 & 5e-4 & 8e-4 & 1e-3 & 3e-3 & 8e-3 & 2e-2 \\
    \hline
     DLGI \cite{wang2019learning}  & 14.54$\pm$1.588 & 14.38$\pm$1.536 &  14.17$\pm$1.531 &  13.98$\pm$1.425 & 13.51$\pm$1.541 & 12.87$\pm$1.390 & 12.36$\pm$1.225 & 11.10$\pm$1.124 \\
     Deepghost \cite{rizvi2020deepghost}  & 17.67$\pm$1.725 & 17.43$\pm$1.577 &  17.32$\pm$1.509 &  17.08$\pm$1.628 & 16.87$\pm$1.486 & 16.44$\pm$1.275 & 15.98$\pm$1.486 & 14.52$\pm$1.043 \\
     SPI-GAN  & \textbf{21.03$\pm$1.789} &  \textbf{20.91$\pm$1.832}   &  \textbf{20.86$\pm$1.810} &  \textbf{20.80$\pm$1.744} & \textbf{20.75$\pm$1.692} & \textbf{20.60$\pm$1.663} & \textbf{19.39$\pm$1.514} & \textbf{17.47$\pm$1.352} \\
\hline
  \end{tabular}

    \label{tbl:PSNR_noise}
\end{table*}

\begin{table*}
    \caption{Average SSIM of different deep learning based methods under different noise levels.}
  \centering
  \begin{tabular}{c|c|c|c|c|c|c|c|c}
    Noise Level & 1e-4 & 3e-4 & 5e-4 & 8e-4 & 1e-3 & 3e-3 & 8e-3 & 2e-2 \\
    \hline
     DLGI \cite{wang2019learning}  & 0.459$\pm$0.094 & 0.450$\pm$0.092 &  0.443$\pm$0.084 &  0.436$\pm$0.087 & 0.431$\pm$0.075 & 0.427$\pm$0.072 & 0.403$\pm$0.081 & 0.332$\pm$0.071 \\

     Deepghost \cite{rizvi2020deepghost}  & 0.531$\pm$0.087 & 0.518$\pm$0.0.079 &  0.491$\pm$0.074 &  0.483$\pm$0.073 & 0.473$\pm$0.068 & 0.462$\pm$0.064 & 0.442$\pm$0.078 & 0.361$\pm$0.061 \\

     SPI-GAN  & \textbf{0.670$\pm$0.072} &  \textbf{0.667$\pm$0.075}   &  \textbf{0.662$\pm$0.068} &  \textbf{0.659$\pm$0.073} & \textbf{0.653$\pm$0.071} & \textbf{0.646$\pm$0.070} & \textbf{0.580$\pm$0.067} & \textbf{0.464$\pm$0.065} \\
    \hline
  \end{tabular}

    \label{tbl:SSIM_noise}
\end{table*}

In addition to these iterative methods, we also consider DL-based methods such as DLGI \cite{wang2019learning}, GIDL \cite{lyu2017deep}, and deepghost \cite{rizvi2020deepghost}. Among these, DLGI and GIDL deal with simple images such as digit images from MNIST. Furthermore, GIDL does not employ any convolution layers and instead reconstructs images using only fully-connected (FC) layers. This limits its reconstruction capability resulting in poor PSNR and SSIM. Unlike convolution layer, FC layer does not benefit from the information such as spatial correlation and structure of various objects in the image. 
For a complex dataset like STL10, this method completely fails to reconstruct the scene. Even for very high number of measurements, the performance of GIDL is poor. Table \ref{tbl:PSNR} is a clear indicator for that. For example, a sampling rate of 25 $\%$ results in an average PSNR of 11.60 dB with 0.169 SSIM. Furthermore, it can be observed that all of the iterative methods outperform GIDL, which is not desirable. 

DLGI tries to perform the reconstruction task from linear measurements. This method employs a parallel network architecture with several upsampling layers to increase the resolution of the image. However, these types of layers slow down the training process. Moreover, DLGI is suited for recovering simple images like handwritten digits but fails to reconstruct images that has complex scene in it. 
Another DL-based method, deepghost, is based on the auto-encoder architecture that compresses the image into a lower dimension. This network works as a denoiser that recovers images from their undersampled noisy reconstructions. For this noisy reconstruction, the differential ghost imaging (DGI) technique is being employed. Due to the architectural setup of the network, deepghost fails to learn useful feature representations from large number of noisy data. As the generalization performance mostly depends on the features we learn from the training inputs, the reconstruction performance is not on par with our method. Table \ref{tbl:PSNR} shows that SPI-GAN outperforms deepghost and DLGI in terms average PSNR and SSIM. For example, we obtain an average of $20.32 dB$ PSNR and $0.627$ SSIM when the sampling rate is set to 0.20. Where as, deepghost reconstructs samples with an average PSNR of $17.35 dB$ under the same settings. The performance of DLGI is worse than deepghost, averaging only $14.46 dB$ of PSNR and $0.395$ of SSIM. The performance deteriorates further when we decrease the sampling rate. These two methods consistently perform worse than our method. We also put the qualitative comparison of these methods in the supplementary section. 

\begin{table*}
    \caption{Average PSNR (dB) of different deep learning based methods. We train the network on STL10 training images and test it on images from 6 other datasets. The superior performance compared to other methods proves high generalizabality of our method to completely unseen images.}
  \centering
  \begin{tabular}{c|c|c|c|c|c|c}
    
    Dataset & CBSD68 & Mandrill & Urban-100 & Set14 & SunHays-80 & BSDS300 \\
    \hline
     DLGI \cite{wang2019learning}  & 12.29$\pm$1.623 & 12.17$\pm$1.341 &  11.87$\pm$1.159 &  12.56$\pm$1.073 & 11.98$\pm$1.523 & 11.74$\pm$1.217 \\
     Deepghost \cite{rizvi2020deepghost}  & 14.63$\pm$1.359 & 14.36$\pm$1.855 &  14.21$\pm$1.365 &  14.26$\pm$0.990 & 14.84$\pm$1.127 & 14.06$\pm$1.411 \\
     SPI-GAN  & \textbf{18.28$\pm$1.782} &  \textbf{18.43$\pm$1.695}   &  \textbf{17.15$\pm$1.860} &  \textbf{17.58$\pm$1.052} & \textbf{17.85$\pm$1.567} & \textbf{18.49$\pm$1.666} \\
  \hline
  \end{tabular}

    \label{tbl:PSNR_other_datasets}
\end{table*}

\begin{table*}
    \caption{Average SSIM gain for different unseen datasets.}
  \centering
  \begin{tabular}{c|c|c|c|c|c|c}
    Dataset & CBSD68 & Mandrill & Urban-100 & Set14 & SunHays-80 & BSDS300 \\
    \hline
     DLGI \cite{wang2019learning}  & 0.294$\pm$0.061 & 0.313$\pm$0.097 &  0.308$\pm$0.076 &  0.337$\pm$0.065 & 0.326$\pm$0.064 & 0.319$\pm$0.058 \\

     Deepghost \cite{rizvi2020deepghost}  & 0.451$\pm$0.087 & 0.506$\pm$0.142 &  0.427$\pm$0.090 &  0.515$\pm$0.085 & 0.482$\pm$0.078 & 0.470$\pm$0.093 \\

     SPI-GAN  & \textbf{0.528$\pm$0.074} &  \textbf{0.562$\pm$0.088}   &  \textbf{0.491$\pm$0.072} &  \textbf{0.573$\pm$0.0718} & \textbf{0.512$\pm$0.065} & \textbf{0.553$\pm$0.070} \\
    \hline
  \end{tabular}
    \label{tbl:SSIM_other_datasets}
\end{table*}
\begin{figure*}[htb]
    \centering
    \includegraphics[width=1\linewidth]{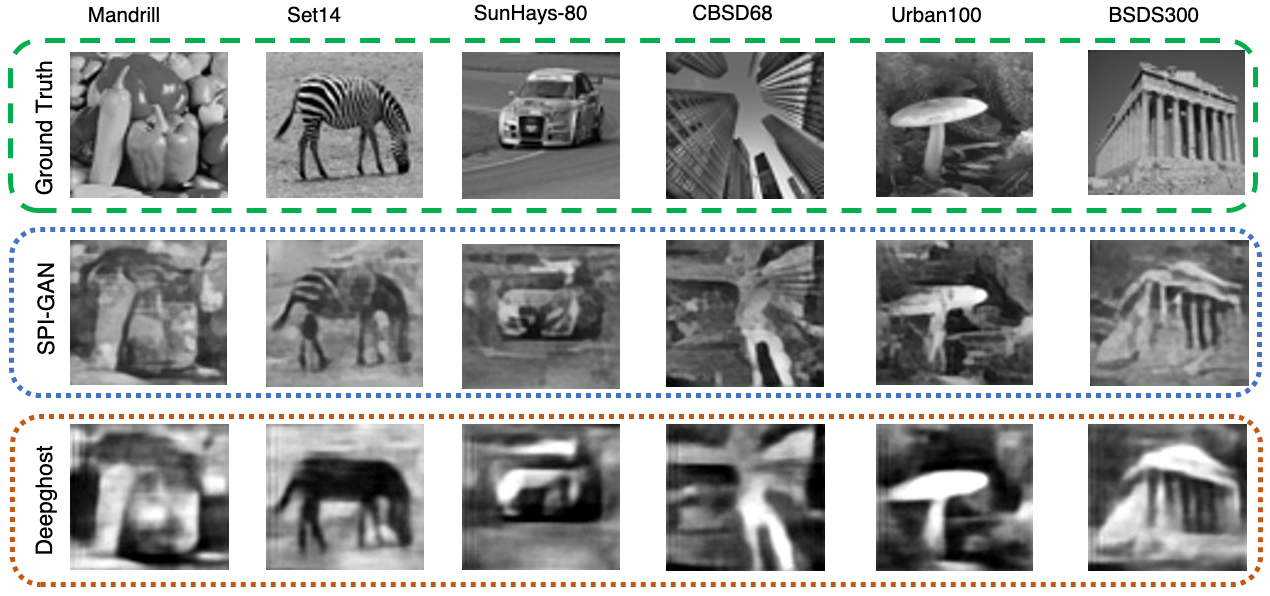}
    \caption{Performance of SPI-GAN on completely 6 unseen datasets. To show the generalization ability, we train our network on STL10 dataset and test it on other datasets. Compared to Deepghost, our method can reconstruct better quality images. Sampling rate is set at 20 $\%$.}
    \label{fig:Other_dataset}
\end{figure*}
\subsection{Measurement Noise}
In general, measurements $\boldsymbol{y}$ get corrupted due to noise derived from different sources such as ambient light, circuit current, etc. Furthermore, single-pixel imaging requires physical devices like SLM and photo detector which could be other sources of noise. Considering all these scenarios, we add a Gaussian noise to the measurements to better represent the real-world setup of SPI. In this section, we analyze the robustness of our method against different levels of noise.  
The Gaussian noise term can be expressed as
\begin{equation}
    \boldsymbol{q}(n) = \frac{1}{\sqrt{2\pi}\sigma} \exp{(- \frac{n^2}{2 \sigma^2})},
\end{equation}
where $n$ is the noise term and $\sigma$ stands for the standard deviation. By varying $\sigma$, we can achieve different noise levels which is added to the measurements. The noise level can be defined as the ratio of $\sigma$ and the number of pixels in the image. 

In our work, we present the reconstruction performance of SPI-GAN under 8 different noise levels. The wide range of noise levels is chosen to show the robustness of our method against noise. In general, it is harder to reconstruction the target image from the corrupted measurements. Both iterative and DL-based methods struggle in the presence of noise. In this stage of comparison, we choose DLGI and Deepghost for comparison since they show some level of robustness against noise. Tables \ref{tbl:PSNR_noise} and \ref{tbl:SSIM_noise} show the average PSNR and SSIM under different noise levels. For a noise level of $1e{-4}$, we achieve a 21.03 dB PSNR compared to 17.67 dB and 14.54 dB for Deepghost and DLGI, respectively. With noise, the SSIM value drops to 0.670 from 0.673 without noise. For noise range of $1e{-4}$ to $3e{-3}$, SPI-GAN, along with other methods, experiences a steady decrease in performance. Even though the performance takes a major hit when the noise level rises to $2e{-2}$, SPI-GAN performs significantly better than other methods. We discuss more on this in the supplementary material.   
\begin{figure*}
    \centering
    \includegraphics[width=1\linewidth]{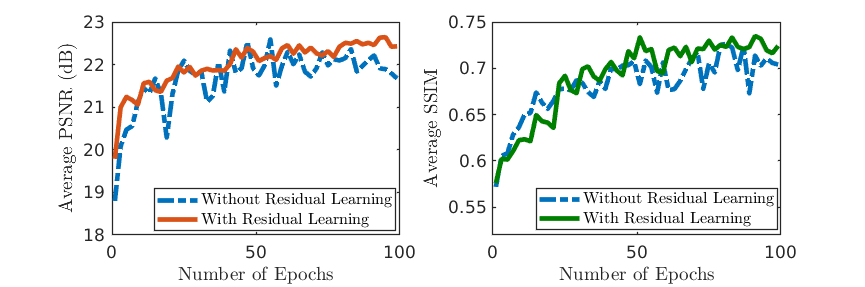}
    \caption{Performance of our network under different network settings. Best performance is obtained when we employ residual learning. All the values are calculated for validation images. Sampling rate is 30 $\%$.}
    \label{fig:residual_final}
\end{figure*}

\subsection{Effect of Residual Learning}
It is well known that deep architectures have better representation capability compared to shallow architectures. However, in many scenarios, deep neural networks are hard to train because of the so called vanishing gradient problem \cite{glorot2010understanding}. To overcome this, architectures with residual blocks were proposed that are resistant to this type of problem \cite{he2016deep}. It has been shown in \cite{zaeemzadeh2020norm} that this type of architecture preserves the norm of the gradient to close to unity. This allows us to use deep network architectures in setting up the GAN framework. Previously, residual learning has been applied for noise removal \cite{song2019dynamic} and image super resolution \cite{Yang_2017,li2018multi,zhang2018residual}. In this section, we highlight the importance of residual learning in image recovery. The experimental results with and without skip connection indicates its impact on the performance of SPI-GAN. Even though we present this case for 30$\%$ sampling ratio, same level of performance gain is achieved for other ratios too. Fig. \ref{fig:residual_final} shows that one can obtain slightly better performance with the help of skip connections. With residual learning, the performance gain is almost 0.6 dB higher for PSNR and 0.025 higher for SSIM. The reason for this bump can be pointed to the better loss gradient flow at the time of training. This in turn helps the network to learn useful features from the data.

\subsection{Proof of Generalizability}
The effectiveness of our work extends to other unseen datasets too. In this section, we show such generalization ability of our method. We choose 6 different widely used datasets in our work. Here, the sampling ratio is fixed at 0.25.

Among them, CBSD68 contains 68 images and BSDS300 \cite{MartinFTM01} is the Berkeley semantic segmentation dataset with 300 images. Table \ref{tbl:PSNR_other_datasets} shows that our method obtains an average PSNR of 18.28 dB and 18.49 dB for these two datasets, respectively.
For comparison, performance of DLGI and Deepghost are also presented in the table. It can be seen that SPI-GAN outperforms both of these methods by a significant margin.  
Our method also obtains a higher average value in terms of SSIM. Another dataset we consider is Mandrill which contains 14 images. The average PSNR and SSIM values are 18.43 dB and 0.562, respectively. Similarly, Set14 \cite{zeyde2010single} also contains 14 images and we obtain good results for this one too. The 5th and 6th datasets we consider are Urban100\cite{Huang-CVPR-2015} and SunHays-80 \cite{sun2012super}. These two datasets consist of 100 and 80 images, respectively. SPI-GAN generalizes better than other methods achieving an average PSNR of 17.15 dB for Urban100 and 17.85 dB for SunHays-80. Whereas, Deepghost achieves a 14.21 dB and 14.84dB PSNR, respectively. The shallow autoencoder architecture of Deepghost could be one reason for that as it lacks representation power. For qualitative comparison, we present some of the reconstructed images from each dataset in Fig. \ref{fig:Other_dataset}. The details in our reconstructed images are better than the Deepghost.  
\begin{table}
    \caption{Imaging time, sum of image acquisition and reconstruction time, for different single-pixel imaging methods. For a 15 $\%$ sampling rate, SPI-GAN can reconstruct video at 15 fps.}
  \centering
  \begin{tabular}{c|c|c|c}
    Method & Reconstruction time & Total Imaging time & FPS  \\
    \hline
     Sparse \cite{duarte2008single} & 3.35 & 3.38 & 0.3 \\
    \hline
     AP \cite{liao2014generalized} & 3.54 & 3.357 & 0.28\\
    \hline
     CGD \cite{5666245} & 0.41 & 0.44 & 2.27 \\
     \hline
     SPI-GAN & 0.035 & 0.065 & $\boldsymbol{15.38}$\\
    \hline
  \end{tabular}

    \label{tbl:Imaging_time}
\end{table}
\subsection{Single-Pixel Video}
Unlike iterative-based methods, DL-based methods can reconstruct images much faster. Table \ref{tbl:Imaging_time} shows the imaging time for different non-DL methods and SPI-GAN. At an SR of $15 \%$, we present the reconstruction time for each of these methods along with their total imaging time. Total imaging time consists of image acquisition and reconstruction time. Considering the DMD array with a modulation rate of 20 KHz, we have an acquisition time of 0.03 sec for an SR of $15\%$. For a fixed modulation rate, the image acquisition time remains the same irrespective of the reconstruction method one employs. The reconstruction time varies depending on the nature of recovery. Usually, iterative methods have higher reconstruction time in contrast to DL-based methods. As shown in Table \ref{tbl:Imaging_time}, methods like AP \cite{liao2014generalized} and Sparse \cite{duarte2008single} have imaging time more than 3 seconds making them inapplicable to video applications. CGD \cite{5666245} offers faster imaging but can recover only 2 image frames per second which is too low for a video. On the other hand, SPI-GAN can reconstruct images much faster compared to these methods. The smaller imaging time offers reconstruction at higher FPS. Depending the sampling rate, our method can reconstruct video with 15 fps (for 15 $\%$ SR) to 22 fps (5 $\%$ SR).  After training for 100 epochs, we obtain an average PSNR and SSIM of 18.95 dB and 0.545 on the validation set, respectively. We attach several original and reconstructed video clips with our supplementary materials.


\section{Conclusion}\label{sec:conclusion}
We have proposed a Generative Adversarial Network (GAN)-based reconstruction framework, refereed to as SPI-GAN, for single-pixel imaging. We demonstrate that by designing proper architectures and loss functions, one can achieve a high performance gain over current iterative and deep learning (DL)-based  methods. We employed a ResNet-like architecture for the generator since it offers certain advantages due to its residual connections. In addition to the commonly used mean squared error (MSE) loss, we employ a perceptual similarity loss to produce a clear reconstruction. Another contribution of our work can be attributed to the single-pixel video reconstruction. Due to better representation learning ability of the generator, SPI-GAN can recover images even from very small number of measurements. This enables us to recover video frames at a much higher rate. Our experimental analysis shows that better representations also allow us to achieve noise robustness. Furthermore, compared to other DL-based methods, our method shows better generalization ability to completely unseen images. In future, SPI-GAN can be extended to reconstruct 3D single-pixel video.

\label{sec:ex}
 
\medskip
\footnotesize
\bibliographystyle{IEEEtran}
\bibliography{IEEEabrv,reference.bib}

\begin{thebibliography}{10}
\providecommand{\url}[1]{#1}
\csname url@samestyle\endcsname
\providecommand{\newblock}{\relax}
\providecommand{\bibinfo}[2]{#2}
\providecommand{\BIBentrySTDinterwordspacing}{\spaceskip=0pt\relax}
\providecommand{\BIBentryALTinterwordstretchfactor}{4}
\providecommand{\BIBentryALTinterwordspacing}{\spaceskip=\fontdimen2\font plus
\BIBentryALTinterwordstretchfactor\fontdimen3\font minus
  \fontdimen4\font\relax}
\providecommand{\BIBforeignlanguage}[2]{{%
\expandafter\ifx\csname l@#1\endcsname\relax
\typeout{** WARNING: IEEEtran.bst: No hyphenation pattern has been}%
\typeout{** loaded for the language `#1'. Using the pattern for}%
\typeout{** the default language instead.}%
\else
\language=\csname l@#1\endcsname
\fi
#2}}
\providecommand{\BIBdecl}{\relax}
\BIBdecl

\bibitem{duarte2008single}
M.~F. Duarte, M.~A. Davenport, D.~Takhar, J.~N. Laska, T.~Sun, K.~F. Kelly, and
  R.~G. Baraniuk, ``Single-pixel imaging via compressive sampling,'' \emph{IEEE
  signal processing magazine}, vol.~25, no.~2, pp. 83--91, 2008.

\bibitem{donoho2006compressed}
D.~L. Donoho, ``Compressed sensing,'' \emph{IEEE Transactions on information
  theory}, vol.~52, no.~4, pp. 1289--1306, 2006.

\bibitem{li2017efficient}
Z.~Li, J.~Suo, X.~Hu, C.~Deng, J.~Fan, and Q.~Dai, ``Efficient single-pixel
  multispectral imaging via non-mechanical spatio-spectral modulation,''
  \emph{Scientific Reports}, vol.~7, no.~1, pp. 1--7, 2017.

\bibitem{wang2015hyperspectral}
Y.~Wang, J.~Suo, J.~Fan, and Q.~Dai, ``Hyperspectral computational ghost
  imaging via temporal multiplexing,'' \emph{IEEE Photonics Technology
  Letters}, vol.~28, no.~3, pp. 288--291, 2015.

\bibitem{bian2016multispectral}
L.~Bian, J.~Suo, G.~Situ, Z.~Li, J.~Fan, F.~Chen, and Q.~Dai, ``Multispectral
  imaging using a single bucket detector,'' \emph{Scientific reports}, vol.~6,
  no.~1, pp. 1--7, 2016.

\bibitem{clemente2010optical}
P.~Clemente, V.~Dur{\'a}n, E.~Tajahuerce, J.~Lancis \emph{et~al.}, ``Optical
  encryption based on computational ghost imaging,'' \emph{Optics letters},
  vol.~35, no.~14, pp. 2391--2393, 2010.

\bibitem{chen2013ghost}
W.~Chen and X.~Chen, ``Ghost imaging for three-dimensional optical security,''
  \emph{Applied Physics Letters}, vol. 103, no.~22, p. 221106, 2013.

\bibitem{zhao2012ghost}
C.~Zhao, W.~Gong, M.~Chen, E.~Li, H.~Wang, W.~Xu, and S.~Han, ``Ghost imaging
  lidar via sparsity constraints,'' \emph{Applied Physics Letters}, vol. 101,
  no.~14, p. 141123, 2012.

\bibitem{gong2016three}
W.~Gong, C.~Zhao, H.~Yu, M.~Chen, W.~Xu, and S.~Han, ``Three-dimensional ghost
  imaging lidar via sparsity constraint,'' \emph{Scientific reports}, vol.~6,
  no.~1, pp. 1--6, 2016.

\bibitem{sun20133d}
B.~Sun, M.~P. Edgar, R.~Bowman, L.~E. Vittert, S.~Welsh, A.~Bowman, and M.~J.
  Padgett, ``3d computational imaging with single-pixel detectors,''
  \emph{Science}, vol. 340, no. 6134, pp. 844--847, 2013.

\bibitem{sun2016single}
M.-J. Sun, M.~P. Edgar, G.~M. Gibson, B.~Sun, N.~Radwell, R.~Lamb, and M.~J.
  Padgett, ``Single-pixel three-dimensional imaging with time-based depth
  resolution,'' \emph{Nature communications}, vol.~7, no.~1, pp. 1--6, 2016.

\bibitem{cheng2009ghost}
J.~Cheng, ``Ghost imaging through turbulent atmosphere,'' \emph{Optics
  express}, vol.~17, no.~10, pp. 7916--7921, 2009.

\bibitem{zhang2010correlated}
P.~Zhang, W.~Gong, X.~Shen, and S.~Han, ``Correlated imaging through
  atmospheric turbulence,'' \emph{Physical Review A}, vol.~82, no.~3, p.
  033817, 2010.

\bibitem{magana2013compressive}
O.~S. Magana-Loaiza, G.~A. Howland, M.~Malik, J.~C. Howell, and R.~W. Boyd,
  ``Compressive object tracking using entangled photons,'' \emph{Applied
  Physics Letters}, vol. 102, no.~23, p. 231104, 2013.

\bibitem{li2014ghost}
E.~Li, Z.~Bo, M.~Chen, W.~Gong, and S.~Han, ``Ghost imaging of a moving target
  with an unknown constant speed,'' \emph{Applied Physics Letters}, vol. 104,
  no.~25, p. 251120, 2014.

\bibitem{gibson2017real}
G.~M. Gibson, B.~Sun, M.~P. Edgar, D.~B. Phillips, N.~Hempler, G.~T. Maker,
  G.~P. Malcolm, and M.~J. Padgett, ``Real-time imaging of methane gas leaks
  using a single-pixel camera,'' \emph{Optics express}, vol.~25, no.~4, pp.
  2998--3005, 2017.

\bibitem{pittman1995optical}
T.~B. Pittman, Y.~Shih, D.~Strekalov, and A.~V. Sergienko, ``Optical imaging by
  means of two-photon quantum entanglement,'' \emph{Physical Review A},
  vol.~52, no.~5, p. R3429, 1995.

\bibitem{strekalov1995observation}
D.~Strekalov, A.~Sergienko, D.~Klyshko, and Y.~Shih, ``Observation of
  two-photon “ghost” interference and diffraction,'' \emph{Physical review
  letters}, vol.~74, no.~18, p. 3600, 1995.

\bibitem{ferri2005high}
F.~Ferri, D.~Magatti, A.~Gatti, M.~Bache, E.~Brambilla, and L.~A. Lugiato,
  ``High-resolution ghost image and ghost diffraction experiments with thermal
  light,'' \emph{Physical review letters}, vol.~94, no.~18, p. 183602, 2005.

\bibitem{bennink2002two}
R.~S. Bennink, S.~J. Bentley, and R.~W. Boyd, ``“two-photon” coincidence
  imaging with a classical source,'' \emph{Physical review letters}, vol.~89,
  no.~11, p. 113601, 2002.

\bibitem{shapiro2008computational}
J.~H. Shapiro, ``Computational ghost imaging,'' \emph{Physical Review A},
  vol.~78, no.~6, p. 061802, 2008.

\bibitem{wang2017high}
Y.~Wang, Y.~Liu, J.~Suo, G.~Situ, C.~Qiao, and Q.~Dai, ``High speed
  computational ghost imaging via spatial sweeping,'' \emph{Scientific
  reports}, vol.~7, p. 45325, 2017.

\bibitem{edgar2015simultaneous}
M.~P. Edgar, G.~M. Gibson, R.~W. Bowman, B.~Sun, N.~Radwell, K.~J. Mitchell,
  S.~S. Welsh, and M.~J. Padgett, ``Simultaneous real-time visible and infrared
  video with single-pixel detectors,'' \emph{Scientific reports}, vol.~5, p.
  10669, 2015.

\bibitem{gong2015high}
W.~Gong and S.~Han, ``High-resolution far-field ghost imaging via sparsity
  constraint,'' \emph{Scientific reports}, vol.~5, no.~1, pp. 1--5, 2015.

\bibitem{hu2015patch}
X.~Hu, J.~Suo, T.~Yue, L.~Bian, and Q.~Dai, ``Patch-primitive driven
  compressive ghost imaging,'' \emph{Optics express}, vol.~23, no.~9, pp.
  11\,092--11\,104, 2015.

\bibitem{yu2014adaptive}
W.-K. Yu, M.-F. Li, X.-R. Yao, X.-F. Liu, L.-A. Wu, and G.-J. Zhai, ``Adaptive
  compressive ghost imaging based on wavelet trees and sparse representation,''
  \emph{Optics express}, vol.~22, no.~6, pp. 7133--7144, 2014.

\bibitem{tropp2007signal}
J.~A. Tropp and A.~C. Gilbert, ``Signal recovery from random measurements via
  orthogonal matching pursuit,'' \emph{IEEE Transactions on information
  theory}, vol.~53, no.~12, pp. 4655--4666, 2007.

\bibitem{candes2006stable}
E.~J. Candes, J.~K. Romberg, and T.~Tao, ``Stable signal recovery from
  incomplete and inaccurate measurements,'' \emph{Communications on Pure and
  Applied Mathematics: A Journal Issued by the Courant Institute of
  Mathematical Sciences}, vol.~59, no.~8, pp. 1207--1223, 2006.

\bibitem{yang2010fast}
A.~Y. Yang, S.~S. Sastry, A.~Ganesh, and Y.~Ma, ``Fast l1-minimization
  algorithms and an application in robust face recognition: A review,'' in
  \emph{2010 IEEE international conference on image processing}.\hskip 1em plus
  0.5em minus 0.4em\relax IEEE, 2010, pp. 1849--1852.

\bibitem{ferri2010differential}
F.~Ferri, D.~Magatti, L.~Lugiato, and A.~Gatti, ``Differential ghost imaging,''
  \emph{Physical review letters}, vol. 104, no.~25, p. 253603, 2010.

\bibitem{gong2010method}
W.~Gong and S.~Han, ``A method to improve the visibility of ghost images
  obtained by thermal light,'' \emph{Physics Letters A}, vol. 374, no.~8, pp.
  1005--1008, 2010.

\bibitem{shin2016performance}
D.~Shin, J.~H. Shapiro, and V.~K. Goyal, ``Performance analysis of low-flux
  least-squares single-pixel imaging,'' \emph{IEEE Signal Processing Letters},
  vol.~23, no.~12, pp. 1756--1760, 2016.

\bibitem{5666245}
N.~{Wang} and Y.~{Wang}, ``An image reconstruction algorithm based on
  compressed sensing using conjugate gradient,'' in \emph{2010 4th
  International Universal Communication Symposium}, 2010, pp. 374--377.

\bibitem{liao2014generalized}
X.~Liao, H.~Li, and L.~Carin, ``Generalized alternating projection for
  weighted-2,1 minimization with applications to model-based compressive
  sensing,'' \emph{SIAM Journal on Imaging Sciences}, vol.~7, no.~2, pp.
  797--823, 2014.

\bibitem{abetamann2013compressive}
M.~A$\beta$mann and M.~Bayer, ``Compressive adaptive computational ghost
  imaging,'' \emph{Scientific reports}, vol.~3, no.~1, pp. 1--5, 2013.

\bibitem{suo2016signal}
J.~Suo, L.~Bian, F.~Chen, and Q.~Dai, ``Signal-dependent noise removal for
  color videos using temporal and cross-channel priors,'' \emph{Journal of
  Visual Communication and Image Representation}, vol.~36, pp. 130--141, 2016.

\bibitem{quantic}
D.~M. Fletcher, ``{QuantIC Business Development Manager},''
  \url{https://quantic.ac.uk/quantic/wp-content/uploads/2016/10/Single-Pixel-Camera-Flyer_FINAL_WEB.pdf},
  2016, [Online; Flyer of single pixel camera].

\bibitem{lyu2017deep}
M.~Lyu, W.~Wang, H.~Wang, H.~Wang, G.~Li, N.~Chen, and G.~Situ,
  ``Deep-learning-based ghost imaging,'' \emph{Scientific reports}, vol.~7,
  no.~1, pp. 1--6, 2017.

\bibitem{wang2019learning}
F.~Wang, H.~Wang, H.~Wang, G.~Li, and G.~Situ, ``Learning from simulation: An
  end-to-end deep-learning approach for computational ghost imaging,''
  \emph{Optics express}, vol.~27, no.~18, pp. 25\,560--25\,572, 2019.

\bibitem{rizvi2020deepghost}
S.~Rizvi, J.~Cao, K.~Zhang, and Q.~Hao, ``Deepghost: real-time computational
  ghost imaging via deep learning,'' \emph{Scientific Reports}, vol.~10, no.~1,
  pp. 1--9, 2020.

\bibitem{yang2019deep}
W.~Yang, X.~Zhang, Y.~Tian, W.~Wang, J.-H. Xue, and Q.~Liao, ``Deep learning
  for single image super-resolution: A brief review,'' \emph{IEEE Transactions
  on Multimedia}, vol.~21, no.~12, pp. 3106--3121, 2019.

\bibitem{ledig2017photo}
C.~Ledig, L.~Theis, F.~Husz{\'a}r, J.~Caballero, A.~Cunningham, A.~Acosta,
  A.~Aitken, A.~Tejani, J.~Totz, Z.~Wang \emph{et~al.}, ``Photo-realistic
  single image super-resolution using a generative adversarial network,'' in
  \emph{Proceedings of the IEEE conference on computer vision and pattern
  recognition}, 2017, pp. 4681--4690.

\bibitem{coates2011analysis}
A.~Coates, A.~Ng, and H.~Lee, ``An analysis of single-layer networks in
  unsupervised feature learning,'' in \emph{Proceedings of the fourteenth
  international conference on artificial intelligence and statistics}.\hskip
  1em plus 0.5em minus 0.4em\relax JMLR Workshop and Conference Proceedings,
  2011, pp. 215--223.

\bibitem{4016283}
E.~J. {Candes} and T.~{Tao}, ``Near-optimal signal recovery from random
  projections: Universal encoding strategies?'' \emph{IEEE Transactions on
  Information Theory}, vol.~52, no.~12, pp. 5406--5425, 2006.

\bibitem{ren2018learning}
Z.~Ren, Z.~Xu, and E.~Y. Lam, ``Learning-based nonparametric autofocusing for
  digital holography,'' \emph{Optica}, vol.~5, no.~4, pp. 337--344, 2018.

\bibitem{wang2018eholonet}
H.~Wang, M.~Lyu, and G.~Situ, ``eholonet: a learning-based end-to-end approach
  for in-line digital holographic reconstruction,'' \emph{Optics express},
  vol.~26, no.~18, pp. 22\,603--22\,614, 2018.

\bibitem{rivenson2018phase}
Y.~Rivenson, Y.~Zhang, H.~G{\"u}nayd{\i}n, D.~Teng, and A.~Ozcan, ``Phase
  recovery and holographic image reconstruction using deep learning in neural
  networks,'' \emph{Light: Science \& Applications}, vol.~7, no.~2, pp.
  17\,141--17\,141, 2018.

\bibitem{lyu2019learning}
M.~Lyu, H.~Wang, G.~Li, S.~Zheng, and G.~Situ, ``Learning-based lensless
  imaging through optically thick scattering media,'' \emph{Advanced
  Photonics}, vol.~1, no.~3, p. 036002, 2019.

\bibitem{li2018deep}
Y.~Li, Y.~Xue, and L.~Tian, ``Deep speckle correlation: a deep learning
  approach toward scalable imaging through scattering media,'' \emph{Optica},
  vol.~5, no.~10, pp. 1181--1190, 2018.

\bibitem{li2018imaging}
S.~Li, M.~Deng, J.~Lee, A.~Sinha, and G.~Barbastathis, ``Imaging through glass
  diffusers using densely connected convolutional networks,'' \emph{Optica},
  vol.~5, no.~7, pp. 803--813, 2018.

\bibitem{sinha2017lensless}
A.~Sinha, J.~Lee, S.~Li, and G.~Barbastathis, ``Lensless computational imaging
  through deep learning,'' \emph{Optica}, vol.~4, no.~9, pp. 1117--1125, 2017.

\bibitem{wu2016artificial}
G.~Wu, T.~Nowotny, Y.~Zhang, H.-Q. Yu, and D.~D.-U. Li, ``Artificial neural
  network approaches for fluorescence lifetime imaging techniques,''
  \emph{Optics letters}, vol.~41, no.~11, pp. 2561--2564, 2016.

\bibitem{he2018ghost}
Y.~He, G.~Wang, G.~Dong, S.~Zhu, H.~Chen, A.~Zhang, and Z.~Xu, ``Ghost imaging
  based on deep learning,'' \emph{Scientific reports}, vol.~8, no.~1, pp. 1--7,
  2018.

\bibitem{goodfellow2014generative}
I.~J. Goodfellow, J.~Pouget-Abadie, M.~Mirza, B.~Xu, D.~Warde-Farley, S.~Ozair,
  A.~Courville, and Y.~Bengio, ``Generative adversarial networks,'' \emph{arXiv
  preprint arXiv:1406.2661}, 2014.

\bibitem{zhu2017unpaired}
J.-Y. Zhu, T.~Park, P.~Isola, and A.~A. Efros, ``Unpaired image-to-image
  translation using cycle-consistent adversarial networks,'' in
  \emph{Proceedings of the IEEE international conference on computer vision},
  2017, pp. 2223--2232.

\bibitem{isola2017image}
P.~Isola, J.-Y. Zhu, T.~Zhou, and A.~A. Efros, ``Image-to-image translation
  with conditional adversarial networks,'' in \emph{Proceedings of the IEEE
  conference on computer vision and pattern recognition}, 2017, pp. 1125--1134.

\bibitem{mirza2014conditional}
M.~Mirza and S.~Osindero, ``Conditional generative adversarial nets,''
  \emph{arXiv preprint arXiv:1411.1784}, 2014.

\bibitem{rajeswar2017adversarial}
S.~Rajeswar, S.~Subramanian, F.~Dutil, C.~Pal, and A.~Courville, ``Adversarial
  generation of natural language,'' \emph{arXiv preprint arXiv:1705.10929},
  2017.

\bibitem{press2017language}
O.~Press, A.~Bar, B.~Bogin, J.~Berant, and L.~Wolf, ``Language generation with
  recurrent generative adversarial networks without pre-training,'' \emph{arXiv
  preprint arXiv:1706.01399}, 2017.

\bibitem{seeliger2018generative}
K.~Seeliger, U.~G{\"u}{\c{c}}l{\"u}, L.~Ambrogioni,
  Y.~G{\"u}{\c{c}}l{\"u}t{\"u}rk, and M.~A. van Gerven, ``Generative
  adversarial networks for reconstructing natural images from brain activity,''
  \emph{NeuroImage}, vol. 181, pp. 775--785, 2018.

\bibitem{shen2019deep}
G.~Shen, T.~Horikawa, K.~Majima, and Y.~Kamitani, ``Deep image reconstruction
  from human brain activity,'' \emph{PLoS computational biology}, vol.~15,
  no.~1, p. e1006633, 2019.

\bibitem{shen2019end}
G.~Shen, K.~Dwivedi, K.~Majima, T.~Horikawa, and Y.~Kamitani, ``End-to-end deep
  image reconstruction from human brain activity,'' \emph{Frontiers in
  computational neuroscience}, vol.~13, p.~21, 2019.

\bibitem{vanrullen2019reconstructing}
R.~VanRullen and L.~Reddy, ``Reconstructing faces from fmri patterns using deep
  generative neural networks,'' \emph{Communications biology}, vol.~2, no.~1,
  pp. 1--10, 2019.

\bibitem{reed2016generative}
S.~Reed, Z.~Akata, X.~Yan, L.~Logeswaran, B.~Schiele, and H.~Lee, ``Generative
  adversarial text to image synthesis,'' \emph{arXiv preprint
  arXiv:1605.05396}, 2016.

\bibitem{denton2015deep}
E.~L. Denton, S.~Chintala, R.~Fergus \emph{et~al.}, ``Deep generative image
  models using a laplacian pyramid of adversarial networks,'' \emph{Advances in
  neural information processing systems}, vol.~28, pp. 1486--1494, 2015.

\bibitem{sampsell1994digital}
J.~B. Sampsell, ``Digital micromirror device and its application to projection
  displays,'' \emph{Journal of Vacuum Science \& Technology B: Microelectronics
  and Nanometer Structures Processing, Measurement, and Phenomena}, vol.~12,
  no.~6, pp. 3242--3246, 1994.

\bibitem{wold1987principal}
S.~Wold, K.~Esbensen, and P.~Geladi, ``Principal component analysis,''
  \emph{Chemometrics and intelligent laboratory systems}, vol.~2, no. 1-3, pp.
  37--52, 1987.

\bibitem{ahmed1974discrete}
N.~Ahmed, T.~Natarajan, and K.~R. Rao, ``Discrete cosine transform,''
  \emph{IEEE transactions on Computers}, vol. 100, no.~1, pp. 90--93, 1974.

\bibitem{baraniuk2007compressive}
R.~G. Baraniuk, ``Compressive sensing [lecture notes],'' \emph{IEEE signal
  processing magazine}, vol.~24, no.~4, pp. 118--121, 2007.

\bibitem{simonyan2014very}
K.~Simonyan and A.~Zisserman, ``Very deep convolutional networks for
  large-scale image recognition,'' \emph{arXiv preprint arXiv:1409.1556}, 2014.

\bibitem{mathieu2015deep}
M.~Mathieu, C.~Couprie, and Y.~LeCun, ``Deep multi-scale video prediction
  beyond mean square error,'' \emph{arXiv preprint arXiv:1511.05440}, 2015.

\bibitem{johnson2016perceptual}
J.~Johnson, A.~Alahi, and L.~Fei-Fei, ``Perceptual losses for real-time style
  transfer and super-resolution,'' in \emph{European conference on computer
  vision}.\hskip 1em plus 0.5em minus 0.4em\relax Springer, 2016, pp. 694--711.

\bibitem{wang2004image}
Z.~Wang, A.~C. Bovik, H.~R. Sheikh, and E.~P. Simoncelli, ``Image quality
  assessment: from error visibility to structural similarity,'' \emph{IEEE
  transactions on image processing}, vol.~13, no.~4, pp. 600--612, 2004.

\bibitem{1284395}
{Zhou Wang}, A.~C. {Bovik}, H.~R. {Sheikh}, and E.~P. {Simoncelli}, ``Image
  quality assessment: from error visibility to structural similarity,''
  \emph{IEEE Transactions on Image Processing}, vol.~13, no.~4, pp. 600--612,
  2004.

\bibitem{glorot2010understanding}
X.~Glorot and Y.~Bengio, ``Understanding the difficulty of training deep
  feedforward neural networks,'' in \emph{Proceedings of the thirteenth
  international conference on artificial intelligence and statistics}.\hskip
  1em plus 0.5em minus 0.4em\relax JMLR Workshop and Conference Proceedings,
  2010, pp. 249--256.

\bibitem{he2016deep}
K.~He, X.~Zhang, S.~Ren, and J.~Sun, ``Deep residual learning for image
  recognition,'' in \emph{Proceedings of the IEEE conference on computer vision
  and pattern recognition}, 2016, pp. 770--778.

\bibitem{zaeemzadeh2020norm}
A.~Zaeemzadeh, N.~Rahnavard, and M.~Shah, ``Norm-preservation: Why residual
  networks can become extremely deep?'' \emph{IEEE transactions on pattern
  analysis and machine intelligence}, 2020.

\bibitem{song2019dynamic}
Y.~Song, Y.~Zhu, and X.~Du, ``Dynamic residual dense network for image
  denoising,'' \emph{Sensors}, vol.~19, no.~17, p. 3809, 2019.

\bibitem{Yang_2017}
\BIBentryALTinterwordspacing
W.~Yang, J.~Feng, J.~Yang, F.~Zhao, J.~Liu, Z.~Guo, and S.~Yan, ``Deep edge
  guided recurrent residual learning for image super-resolution,'' \emph{IEEE
  Transactions on Image Processing}, vol.~26, no.~12, p. 5895–5907, Dec 2017.
  [Online]. Available: \url{http://dx.doi.org/10.1109/TIP.2017.2750403}
\BIBentrySTDinterwordspacing

\bibitem{li2018multi}
J.~Li, F.~Fang, K.~Mei, and G.~Zhang, ``Multi-scale residual network for image
  super-resolution,'' in \emph{Proceedings of the European Conference on
  Computer Vision (ECCV)}, 2018, pp. 517--532.

\bibitem{zhang2018residual}
Y.~Zhang, Y.~Tian, Y.~Kong, B.~Zhong, and Y.~Fu, ``Residual dense network for
  image super-resolution,'' in \emph{Proceedings of the IEEE conference on
  computer vision and pattern recognition}, 2018, pp. 2472--2481.

\bibitem{MartinFTM01}
D.~Martin, C.~Fowlkes, D.~Tal, and J.~Malik, ``A database of human segmented
  natural images and its application to evaluating segmentation algorithms and
  measuring ecological statistics,'' in \emph{Proc. 8th Int'l Conf. Computer
  Vision}, vol.~2, July 2001, pp. 416--423.

\bibitem{zeyde2010single}
R.~Zeyde, M.~Elad, and M.~Protter, ``On single image scale-up using
  sparse-representations,'' in \emph{International conference on curves and
  surfaces}.\hskip 1em plus 0.5em minus 0.4em\relax Springer, 2010, pp.
  711--730.

\bibitem{Huang-CVPR-2015}
J.-B. Huang, A.~Singh, and N.~Ahuja, ``Single image super-resolution from
  transformed self-exemplars,'' in \emph{Proceedings of the IEEE Conference on
  Computer Vision and Pattern Recognition}, 2015, pp. 5197--5206.

\bibitem{sun2012super}
L.~Sun and J.~Hays, ``Super-resolution from internet-scale scene matching,'' in
  \emph{2012 IEEE International Conference on Computational Photography
  (ICCP)}.\hskip 1em plus 0.5em minus 0.4em\relax IEEE, 2012, pp. 1--12.

\end{thebibliography}
\begin{IEEEbiography}[{\includegraphics[width=1in,height=1in,clip,keepaspectratio]{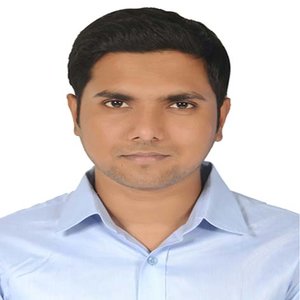}}]{Nazmul Karim}
received his B.S. degree in electrical engineering from the Bangladesh University of Engineering and Technology, Dhaka, Bangladesh, in 2016. He is currently working toward the Ph.D. degree in electrical engineering at the University of Central Florida. His current research interests lie in the areas of machine learning, signal processing, and linear algebra. 
\end{IEEEbiography}

\begin{IEEEbiography}[{\includegraphics[width=1in,height=1.25in,clip,keepaspectratio]{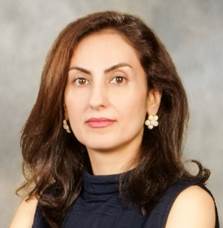}}]{Nazanin Rahnavard}
 (S’97-M’10, SM’19) received her Ph.D. in the School of Electrical and Computer Engineering at the Georgia Institute of Technology, Atlanta, in 2007. She is a Professor in the Department of Electrical and Computer Engineering at the University of Central Florida, Orlando, Florida. Dr. Rahnavard is the recipient of NSF CAREER award in 2011. She has interest and expertise in a variety of research topics in deep learning, communications, networking, and signal processing areas. She serves on the editorial board of the Elsevier Journal on Computer Networks (COMNET) and on the Technical Program Committee of several prestigious international conferences.
\end{IEEEbiography}

\end{document}